%% file: main.tex
\documentclass[10pt,journal,compsoc]{IEEEtran}
\usepackage[T1]{fontenc}
\usepackage{amsmath,amssymb,amsthm}
\usepackage{graphicx,booktabs,multirow}
\usepackage[numbers,sort&compress]{natbib}
\usepackage{enumitem}
\usepackage{xcolor}
\usepackage[hidelinks]{hyperref}
\usepackage{microtype}
\usepackage{placeins}
\usepackage{float}

\usepackage{indentfirst}
\renewcommand{\paragraph}[1]{\par\smallskip\indent\textit{#1.}\ }
\graphicspath{{figs/}}
\newtheorem{definition}{Definition}
\newtheorem{proposition}{Proposition}
\newtheorem{theorem}{Theorem}
\newtheorem{corollary}{Corollary}
\newcommand{\R}{\mathbb{R}}
\newcommand{\DLP}{\mathrm{DLP}}

\begin{document}
\title{Enhancing Local and Global Structure\\ in Compressed AI Agent Memory}
\author{Henry~Han%
\IEEEcompsocitemizethanks{\IEEEcompsocthanksitem H. Han is with the Department of Computer Science, Baylor University, Waco, TX 76798, USA. E-mail: Henry\_Han@baylor.edu}}
\IEEEtitleabstractindextext{%
\input{t_abstract}
\begin{IEEEkeywords}
Vector quantization, embedding compression, agent memory, neighborhood preservation, rank correlation, product quantization, label-free evaluation.
\end{IEEEkeywords}}
\maketitle
\IEEEdisplaynontitleabstractindextext
\IEEEpeerreviewmaketitle
\input{t_intro}
\input{t_related}
\input{t_profile}
\input{t_laq}
\input{t_setup}
\input{t_results_measure}
\input{t_results_laq}
\input{t_discussion}
\bibliographystyle{IEEEtranN}
\bibliography{refs}
\clearpage\onecolumn
\raggedbottom
\makeatletter\setlength{\@fptop}{0pt}\setlength{\@fpbot}{0pt plus 1fil}\makeatother
\renewcommand{\topfraction}{0.95}\renewcommand{\bottomfraction}{0.6}\renewcommand{\textfraction}{0.05}\renewcommand{\floatpagefraction}{0.8}\setcounter{topnumber}{3}\setcounter{totalnumber}{4}
\appendices
\input{t_appendix_proofs}
\input{t_appendix}
\input{t_appendix_d}
\input{t_appendix_runs}
\end{document}

%% file: t_abstract.tex
\begin{abstract}\rightskip=0pt\parfillskip=0pt plus 1fil\relax
AI agents store text as embedding vectors and compress them at scale. A compression is accepted when each item's nearest neighbors survive, but agents also consolidate, delete, and compare memories, tasks that depend on which items lie far apart. We evaluate both ends with two label-free fidelities from our prior work: the local fidelity $L$ at 10 neighbors and the global fidelity $G$ at 30 to 50 percent of the memory. At 96 bytes per passage, rescored binary codes reach an $L$ of 0.97 yet move 10.1 percent of passages to another theme; PCA with 8-bit codes, at an $L$ of 0.77, moves 1.7 percent. Rescoring leaves the stored codes, and thus $G$, unchanged. We prove that across the $G$ band, $1-\nu_k$ is at most 2 to 2.18 times the square root of the near--far rank-violation rate, and that this rate follows the distance noise of the compression (correlation 0.979). Our locality-aware quantization (LAQ) turns the bound into a hinge objective on the codebooks of an OPQ index, keeping its bytes, rotation, and search. To our knowledge, LAQ is the first quantizer whose objective comes from a proven bound on the far end. Reconstruction alone raises $G$ in 0 of 52 held-out runs; the far hinge raises it in all 52. LAQ raises $G$ in all 52 ($p=8.5\times10^{-4}$ over 13 memory--encoder groups), including the external FinDER memory, has the highest $G$ of seven equal-size quantizers in all 12 Multi-Doc-2025 runs, scales to a million passages, and keeps evidence recall.
\end{abstract}

%% file: t_intro.tex
\section{Introduction}\label{sec:intro}

\IEEEPARstart{L}{anguage-model} agents remember by embedding what they read, say, and do. When they need an item, they recall the nearest embeddings~\citep{lewis2020rag}. Managed agent memories extract records from conversations, retrieve them by semantic search, and consolidate them over time~\citep{agentcore2025}. At millions of vectors, every byte per vector costs memory, latency, and money. The vectors are therefore compressed: by truncating Matryoshka embeddings~\citep{kusupati2022matryoshka}, by projection, by scalar or binary codes, or by product quantization~\citep{jegou2011pq}. Cloud services offer each of these as a switch~\citep{titanbinary2024,osquant2025}.

How should an agent memory be compressed so that it keeps both what lies near and what lies far? Today, a compression is accepted by its nearest neighbors. Vendors report that binary embeddings retain 97 to 98.5 percent of retrieval accuracy~\citep{titanbinary2024} and evaluate quantized indexes by recall@100~\citep{osquant2025}. At 96 bytes per passage on FinQA, rescored binary codes keep $L$ at 0.97 and top-10 agreement at 0.87, yet move 10.1 percent of passages to another theme (Section~\ref{sec:measure}). Embedding compression studies report nDCG@10~\citep{huerga2025} or benchmark task scores~\citep{kisako2026}. Yet an agent also acts on what lies far apart: it consolidates a year of filings into themes~\citep{sarthi2024raptor,grootendorst2022bertopic}, deletes every item of a client or a session on request, and maps how companies and sectors relate. Because these operations compare each item with the whole memory, they decide what a sector brief says and what a deletion removes. Keeping this far end is difficult for two reasons: a private agent memory has no labels to test it with, and quantizers are trained for reconstruction, nearest neighbors, or query relevance. Our idea is to measure the far end with the same statistic that measures the near end, prove which pairwise order controls it, and train the codebooks on that order.

\begin{figure*}[p]
\centering
\includegraphics[width=\textwidth,height=0.8\textheight,keepaspectratio]{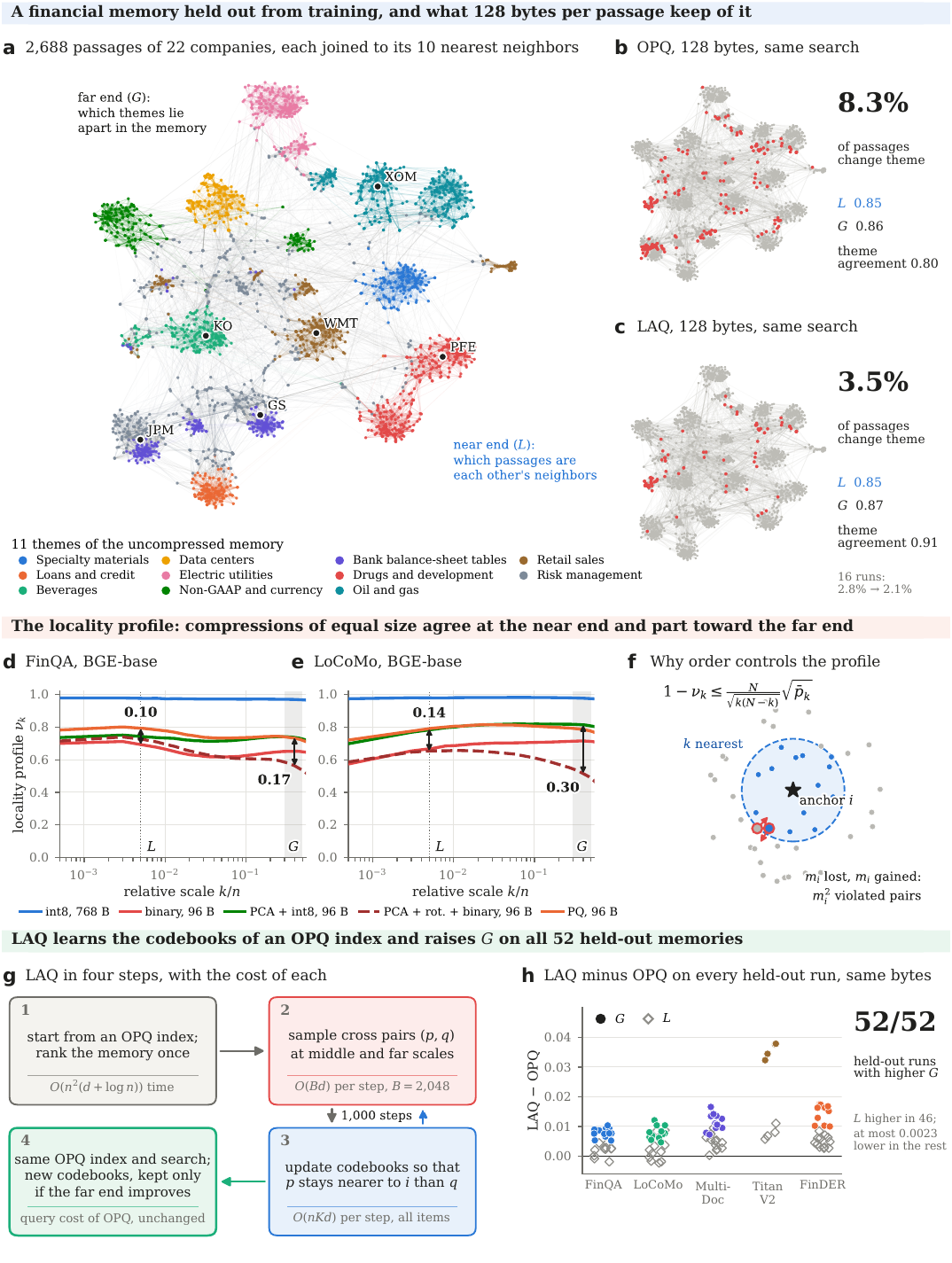}
\caption{At the same bytes and search, locality-aware quantization (LAQ) keeps the near end of an optimized product quantizer (OPQ) and moves fewer passages to another theme. $L$: local fidelity (chance-corrected recall@10); $G$: global fidelity (the profile averaged over 30 to 50 percent of the memory). (a) Multi-Doc-2025~\citep{multidoc2025}, BGE-large~\citep{xiao2024cpack}: 2,688 held-out passages of 22 companies, each joined to its 10 nearest neighbors and colored by theme (t-SNE layout~\citep{vandermaaten2008}); companies form islands, and banks share themes. (b, c) At 128 bytes, both keep $L$ at 0.85; OPQ moves 8.3 percent of the passages to another theme (red) and LAQ 3.5 percent, on the split where OPQ has its lowest theme agreement; over 16 runs the mean falls from 2.8 to 2.1 percent. (d, e) Locality profiles (Definition~\ref{def:profile}) of five compressions of FinQA~\citep{chen2021finqa} and LoCoMo~\citep{maharana2024locomo}: compressions of equal size agree near $k=10$ and part by up to 0.30 in the $G$ band (gray). (f) Every lost--gained neighbor pair is a violated cross pair, which bounds the profile (Theorem~\ref{thm:bound}). (g) LAQ in four steps; gray: cost of each step (Section~\ref{sec:laq}). (h) LAQ minus OPQ on all 52 held-out runs: $G$ rises in all 52 and $L$ in 46.}\label{fig:memgraph}
\end{figure*}

At the same bytes and search, our proposed locality-aware quantization (LAQ) keeps the nearest neighbors of a held-out financial memory while moving far fewer passages to another theme (Figure~\ref{fig:memgraph}a--c). In such a memory, passages from a single company form islands: in the outer fifth of the map, all 10 nearest neighbors of each passage belong to its own company. Passages from different banks meet in shared themes. At 128 bytes per passage, an optimized product quantization (OPQ) index~\citep{ge2014opq} keeps local fidelity $L$ at 0.85 but moves 224 of the 2,688 passages to another theme. LAQ keeps the bytes, the search, and $L$ at 0.85, while moving only 94 passages.

\paragraph{Measurement} One label-free profile reads both ends of a compressed memory. Following our earlier work~\citep{hanLG2026}, we evaluate the two ends of a chance-corrected locality profile~\citep{han2022}: the near end (local structure) and the far end (global structure). Local fidelity $L$ is chance-corrected recall@10. Global fidelity $G$ averages, over 30 to 50 percent of the memory, the agreement on which items lie in each item's nearer part. At its upper edge, the profile measures agreement on the near and far halves, which under a Gaussian copula equals Kendall's $\tau$ of the distances. Both fidelities require only the vectors before and after compression, and $G$ is estimated from a subsample of 2,000 items, however large the memory.

\paragraph{Learning} We prove that rank violations among cross pairs control the profile at every scale. A cross pair consists of one near item and one far item relative to an anchor. Let $\nu_k$ be the profile at scale $k$ and $N=n-1$ for a memory of $n$ items. Then $1-\nu_k$ is at most $N/\sqrt{k(N-k)}$ times the square root of the cross-pair violation rate. The leading factor stays between 2 and 2.18 across the $G$ band. A hinge loss on distance gaps bounds this violation rate. LAQ lowers this hinge at middle and far scales by refining the codebooks of an OPQ index without labels or queries. The principal component analysis (PCA) basis, rotation, bytes per item, and search remain identical to those of the OPQ index.

\paragraph{What is new} To our knowledge, LAQ is the first quantizer trained on the far end of a memory, and its objective comes from a bound we prove. The closest learned quantizers train codebooks with pairwise or triplet losses. PQN uses triplets from class labels~\citep{yu2018pqn}, and spreading vectors pair a nearest-neighbor triplet loss with an entropy term~\citep{sablayrolles2019spreading}. DPQ trains product codes end to end via straight-through assignments~\citep{chen2020dpq}, and JPQ and Distill-VQ train with queries and relevance labels~\citep{zhan2021jpq,xiao2022distillvq}. LAQ differs in target, input, and output. Its target is the rank order at 30 to 50 percent of the memory, selected via our theoretical bound; its input is the unlabeled memory alone; and its output modifies only the codebook values of an OPQ index. We arrived at $\nu_k$ from our DLP curve~\citep{han2022}. Rank-based quality criteria~\citep{lee2009,lee2015} also count kept neighbors across scales and summarize them by an area that weights the near end. To our knowledge, the far end had no proven meaning until our earlier work showed that its midpoint is a rank correlation of distances~\citep{hanLG2026} (Propositions~\ref{prop:duality}--\ref{prop:sheppard}). This paper takes the far end from a measure to a target. We prove that rank violations bound the profile at every scale (Theorem~\ref{thm:bound}), that under noisy order the profile is fixed by one ratio, the distance noise of a compression against the spread of distances (Proposition~\ref{prop:noise}, Corollary~\ref{cor:noiseprofile}), and that at an OPQ start the far hinge falls at no first-order cost in reconstruction (Proposition~\ref{prop:start}). To our knowledge, no earlier quantizer, learned or not, trains its codes on the far end of this curve.

\paragraph{Contributions} Our contributions, in order of importance, are:
\begin{enumerate}[leftmargin=1.6em,itemsep=2pt]
\item \emph{A quantizer derived from a bound.} A new rank-violation bound ties the profile at every scale to the order of cross pairs, and a hinge surrogate makes it trainable. A noisy-order proposition and its corollary predict the violation rate and the profile in the $G$ band from one ratio, the distance noise of the compression against the spread of distances. A further proposition shows that, at the OPQ start, the far hinge falls at no first-order cost in reconstruction, and the straight-through steps of training meet its condition at every start. Each result comes with a proof. LAQ turns the bound into a label-free objective on the codebooks of an OPQ index and replaces that index at the same bytes and search. The ablation confirms that the gain comes from the far hinge: reconstruction alone raises $G$ in 0 of 52 held-out runs, the near hinge raises it by 0.0058 on average, and the far hinge raises it in all 52, by 0.0127.
\item \emph{Held-out evidence that it works.} On held-out companies and conversations across three memories, LAQ raises $G$ in all 40 runs and $L$ in 34. The gain is largest with Amazon Titan V2~\citep{titanbinary2024}, where $G$ rises by 0.032 to 0.038 on every split, two to five times the gain with the three retrieval-tuned encoders. On FinDER~\citep{choi2025finder}, an external financial memory run once after the recipe was fixed, LAQ raises both fidelities in all 12 runs. On Multi-Doc-2025 it lifts the lowest theme agreement from 0.80 to 0.89 and has the highest $G$ among seven quantizers of equal size. Paired tests confirm the gain: $G$ rises in all 13 memory--encoder groups ($p=8.5\times10^{-4}$, sign test), LAQ has a higher $G$ than each of six other quantizers in all 12 Multi-Doc-2025 runs, and evidence recall@10 equals that of OPQ ($p=0.84$). On a million passages, one training run takes 175 seconds and a second run reproduces the codebooks bit for bit.
\item \emph{Why the gain matters.} Each end governs its own operations. Across two agent memories and six encoders, retrieval tracks $L$ (partial rank correlation 0.83 to 0.99 given $G$). With the retrieval-tuned encoders, consolidation and deletion track $G$ (0.55 to 0.93 given $L$). Rescoring repairs the near end at query time but leaves the far end of the stored codes unchanged.
\end{enumerate}

LAQ applies wherever a vector index also serves clustering, deduplication, or mapping.

Section~\ref{sec:related} reviews related work. Section~\ref{sec:profile} presents the measurement theory and the bound, and Section~\ref{sec:laq} introduces the LAQ quantizer. Section~\ref{sec:setup} describes the memories, encoders, and operations. Section~\ref{sec:measure} shows what each end measures, and Section~\ref{sec:laqresults} evaluates LAQ on held-out memories.

%% file: t_related.tex
\section{Related work}\label{sec:related}
We draw on three lines of research and apply them to agent memory: vector quantization, the evaluation of compressed vectors, and local and global structure.

\paragraph{Vector quantization} Quantizers compress vectors through subspace decomposition, additive codewords, or binary projections. Product quantization (PQ) splits a vector into subvectors and codes each separately~\citep{jegou2011pq}, while OPQ learns a rotation to balance variance across subvectors~\citep{ge2014opq}. Residual quantization (RQ)~\citep{chen2010rvq} and local search quantization (LSQ++)~\citep{martinez2018lsq} represent a vector as a sum of codewords, whereas ScaNN weights quantization error by its impact on query inner products~\citep{guo2020scann}. For binary representations, iterative quantization (ITQ) rotates the data before taking signs~\citep{gong2013itq}, and RaBitQ generates randomized binary codes with a theoretical error bound on distance estimates~\citep{gao2024rabitq}. Learned quantizers train codebooks end to end. PQN applies an asymmetric triplet loss on labeled images~\citep{yu2018pqn}. UNQ trains a nonlinear encoder for nearest-neighbor search~\citep{morozov2019unq}. Spreading vectors combine a nearest-neighbor triplet loss with an entropy regularizer~\citep{sablayrolles2019spreading}. DPQ trains product quantization end to end via soft assignments and straight-through estimators~\citep{chen2020dpq}. In dense retrieval, JPQ, RepCONC, and Distill-VQ optimize quantizers using queries and relevance labels or teacher rankings~\citep{zhan2021jpq,zhan2022repconc,xiao2022distillvq}. Unlike these methods, our proposed LAQ trains codebooks for far-end rank order using the unlabeled memory alone.

\paragraph{Evaluation of compressed vectors} Quantized indexes are typically evaluated by nearest-neighbor recall~\citep{aumuller2020annbench,osquant2025}, retrieval quality~\citep{titanbinary2024,huerga2025}, or downstream task scores~\citep{kisako2026}. \citet{kisako2026} conjecture that both clustering and retrieval depend on neighborhood geometry, and \citet{kuffo2026} find that one-bit codes support the internal clustering an index uses to partition its vectors. Earlier, \citet{turaga2009} showed that fitting a moment-preserving quantizer per cluster preserves a known $k$-means partition. In contrast, our measure asks whether the compressed memory keeps both the partition and the global arrangement of the uncompressed memory, with no partition given in advance.

\paragraph{Local and global structure} Rank-based quality criteria evaluate neighborhood preservation across scales~\citep{lee2009,lee2015}. \citet{lueks2011} split the co-ranking curve into local and global regimes, and recent visualization methods such as DREAMS and CMET balance or score both ends~\citep{kury2026,ghosh2025cmet}. Global embedding structure is also commonly scored by pairwise distance correlation and triplet accuracy~\citep{kobak2019,wang2021}. In our prior work, we introduced the degree of locality preservation~\citep{han2022}. In a subsequent study~\citep{hanLG2026}, we defined the local fidelity $L$, global fidelity $G$, and entity fidelity on its chance-corrected profile and proved that the profile's midpoint converges to a rank correlation of distances. We also derived the subsample estimator of $G$ and applied these fidelities to two-dimensional maps and compressed agent memory. Building on that measurement framework, the present paper proves which pairwise order controls the profile, turns the resulting bound into a trainable quantizer, and evaluates both ends on held-out memories.

\paragraph{Agent memory} Benchmarks such as LoCoMo~\citep{maharana2024locomo} and LongMemEval~\citep{wu2025longmemeval} evaluate whether agents can recall facts from long interactions. Agent memory systems consolidate new records against the most similar stored entries~\citep{agentcore2025}, while corpus-level frameworks cluster embeddings into themes and summaries~\citep{sarthi2024raptor,grootendorst2022bertopic,edge2024graphrag}. Whereas these systems compress memory at the text level through summarization and consolidation, we study the compression of the underlying embedding vectors and the memory-wide operations that depend on them.

%% file: t_profile.tex
\section{Two-end fidelity of a compressed memory}\label{sec:profile}

One chance-corrected profile measures both ends of a compressed memory, while the relative order of near and far items controls this profile at every scale. This section defines the profile, proves what its two ends measure, defines the local and global fidelities $L$ and $G$, and proves the bound that LAQ optimizes in Section~\ref{sec:laq}.

\subsection{Setting}

A memory holds $n$ items with vectors $x_1,\dots,x_n\in\R^d$, where $d$ is the embedding dimension. A compression maps each $x_i$ to a stored code. We write $z_i$ for the vector with which the memory computes distances: the decoded vector for scalar, PCA, and product codes, or the $\pm1$ vector for binary codes compared by Hamming distance. All other distances are Euclidean. Let $N=n-1$. For item $i$, let $r^x_i(j)\in\{1,\dots,N\}$ be the rank of item $j\ne i$ among the other items by increasing distance to $x_i$, with ties broken by index, and let $r^z_i(j)$ be the corresponding rank among the compressed vectors. For $1\le k\le N-1$, the $k$-neighborhoods are $\mathcal N^x_k(i)=\{j:r^x_i(j)\le k\}$ and $\mathcal N^z_k(i)=\{j:r^z_i(j)\le k\}$.

\subsection{The locality profile}
At every scale, the profile counts how many neighbors a compression keeps.

\begin{definition}[Locality profile]\label{def:profile}
The degree of locality preservation at scale $k$ is
\begin{equation}\label{eq:dlp}
\DLP_k=\frac{1}{nk}\sum_{i=1}^n\bigl|\mathcal N^x_k(i)\cap\mathcal N^z_k(i)\bigr|,
\end{equation}
and its chance-corrected value is
\begin{equation}\label{eq:nu}
\nu_k=\frac{N\,\DLP_k-k}{N-k}.
\end{equation}
The locality profile is the curve $k\mapsto\nu_k$, read over the relative scale $k/n$.
\end{definition}

We introduced $\DLP_k$ in our earlier work on the explainability of manifold learning~\citep{han2022}. Rank-based quality criteria for dimensionality reduction~\citep{lee2009,lee2015} also count kept neighbors across scales. At a single scale $k$ with the uncompressed memory as reference, $\DLP_k$ is the recall@$k$ used to evaluate approximate nearest-neighbor indexes~\citep{aumuller2020annbench}. Reading this quantity across every scale yields the profile, and the results below say what its two ends measure. Propositions~\ref{prop:chance}--\ref{prop:sheppard} and Definition~\ref{def:LG} are from our earlier work on nonlinear embeddings~\citep{hanLG2026}; we give the proofs so that the paper is self-contained.

The next result gives the chance level and invariance of the profile: a compression that keeps neighbors at random scores $0$, and codes whose distances differ in scale are compared on one statistic.

\begin{proposition}[Chance level and invariance]\label{prop:chance}
(i) If each $\mathcal N^z_k(i)$ is a uniformly random $k$-subset of the other $N$ items, independent of the uncompressed vectors, then $\mathbb E\,\DLP_k=k/N$ and $\mathbb E\,\nu_k=0$. A compression that keeps every ranking has $\nu_k=1$. (ii) $\nu_k$ is invariant under any rotation, reflection, translation, or uniform scaling of the $z_i$, and under any strictly increasing transformation of either distance.
\end{proposition}

All proofs except that of Theorem~\ref{thm:bound} are in Appendix~\ref{app:proofs}. Invariance (ii) allows one statistic to compare Euclidean distances between decoded vectors, Hamming distances between bits, and distances after a learned rotation. All 486 compressions evaluated in Section~\ref{sec:measure} lie above the chance level; the lowest $G$ among them, 0.04, belongs to UMAP~\citep{mcinnes2018} at 16 dimensions (MobileCLIP2~\citep{faghri2025mobileclip2}, FinQA).

The next result shows what the profile reads at its midpoint: for every item, which half of the memory lies near.

\begin{proposition}[Near--far duality]\label{prop:duality}
Let $\mathcal F^x_k(i)$ and $\mathcal F^z_k(i)$ be the complements of $\mathcal N^x_k(i)$ and $\mathcal N^z_k(i)$ among the other $N$ items. Then $|\mathcal F^x_k(i)\cap\mathcal F^z_k(i)|=N-2k+|\mathcal N^x_k(i)\cap\mathcal N^z_k(i)|$. For $k^\star=N/2$ with $N$ even,
\begin{equation}\label{eq:median}
\begin{aligned}
\nu_{k^\star}&=2A-1,\\
A&=\frac{1}{nN}\sum_{i=1}^n\Bigl|\bigl\{j\ne i:\ \mathbb 1[r^x_i(j)\le k^\star]=\mathbb 1[r^z_i(j)\le k^\star]\bigr\}\Bigr|,
\end{aligned}
\end{equation}
where $A$ is the probability that the uncompressed and the compressed memory agree on whether a random other item lies in the nearer or the farther half.
\end{proposition}

While at small $k$ the profile asks whether an item keeps its nearest neighbors, at $k^\star$ it asks which half of the memory is near and which is far for every item. That question is decided by large distances, and thus by the relative positions of themes, sessions, and companies. The median split also acts as a rank correlation of all distances to an item, and under a Gaussian copula the two coincide.

\begin{proposition}[Median split and rank correlation]\label{prop:sheppard}
Fix item $i$ and let $u$ be a uniformly random other item. Suppose that, after strictly increasing transformations, the pair of distances $\bigl(\|x_i-x_u\|,\|z_i-z_u\|\bigr)$ is bivariate normal with correlation $\gamma_i$, that is, the pair has a Gaussian copula. Then for large $n$ the contribution $2A_i-1$ of item $i$ to $\nu_{k^\star}$, with $A_i$ its term in~\eqref{eq:median}, has expectation
\begin{equation}\label{eq:sheppard}
\mathbb E\,(2A_i-1)=\frac{2}{\pi}\arcsin \gamma_i,
\end{equation}
which is Kendall's $\tau$ of the pair and a strictly increasing function of its Spearman correlation $\rho^S_i=\frac{6}{\pi}\arcsin\frac{\gamma_i}{2}$.
\end{proposition}

Compressed memories follow this model closely. Across all 486 compressions, converting the Spearman correlation of all distances into $\frac2\pi\arcsin \gamma$ via Proposition~\ref{prop:sheppard} yields a rank correlation of 0.98 with the measured median split, and the two differ by only 0.03 on average.

\subsection{Local and global fidelity}
Two numbers summarize the profile, one read at each end.

\begin{definition}[Local and global fidelity]\label{def:LG}
The local fidelity and the global fidelity of a compressed memory are
\begin{equation}\label{eq:LG}
L=\nu_{10},\quad G=\frac{1}{|\mathcal K|}\sum_{k\in\mathcal K}\nu_k,\quad \mathcal K=\{k: 0.3\le k/n\le 0.5\}.
\end{equation}
\end{definition}

Here, $L$ is chance-corrected recall@10, whereas $G$ averages the profile over the band whose upper edge is the median split of Proposition~\ref{prop:duality}. Averaging over this band reduces the variance of a single $\nu_{k^\star}$ while preserving the meaning of the median split. Within each of the 18 runs in Section~\ref{sec:measure}, $G$ and the median split have a rank correlation of 0.99 to 1.00 across the 27 compressions. Because both fidelities evaluate the same chance-corrected statistic at two scales, they can be compared directly. Every compression thus lies on the $(L,G)$ plane, where its vertical distance below the diagonal $G=L$ measures how much more of the far end it loses relative to the near end.

Because scale is defined relative to $n$, $G$ can be estimated from a subsample: if $n_0$ items are drawn uniformly, then for a relative scale $\alpha=k/n$ the $\lfloor\alpha n_0\rfloor$ nearest items within the subsample approximate the $\alpha$-quantile neighborhood of the whole memory. We compute $G$ on five uniform subsamples of 2,000 items and report the mean. Across twelve datasets of up to 1.38 million items, subsamples of 2,000 items estimate $G$ within 0.010 of its value at 8,000 items~\citep{hanLG2026}. By contrast, $L$ refers to the fixed scale $k=10$ and is computed on the whole memory. For each pair of items we record the neighbor rank in the uncompressed and in the compressed memory. $\DLP_k$ counts the pairs ranked within $k$ in both, so one running count over $k$ gives $\DLP_k$ and $\nu_k$ at every scale in $O(n_0^2\log n_0)$ time.

\paragraph{Which vectors each fidelity reads} Each fidelity is read from the vectors that its corresponding operations use. When a retrieval pipeline uses rescoring, it fetches the 40 nearest codes and rescores them using full-precision vectors read from disk. Rescoring modifies the lists returned to a query while leaving the stored codes unchanged. We therefore read $L$ and retrieval from the rescored lists seen by the query. We read $G$ and every other operation from the stored codes on which those operations run.

Figure~\ref{fig:memgraph}d, e shows the profiles of real compressions. On the two agent memories, scalar quantization to 8 bits keeps the profile near 1 at every scale. Four compressions to 96 bytes start at similar heights near $k=10$ and separate as $k$ grows. Rotated binary codes fall fastest toward the $G$ band.

\subsection{Rank violations bound the profile}\label{sec:bound}

The order of near and far items controls the profile at every scale. Whereas the profile counts kept neighbors, a learning objective requires a quantity that varies smoothly with the codes. Pairwise rank order provides such a quantity. For item $i$, the anchor, and scale $k$, we call $(p,q)$ a \emph{cross pair} if $p\in\mathcal N^x_k(i)$ and $q\notin\mathcal N^x_k(i)$; that is, $p$ is near and $q$ is far in the uncompressed memory. Item $i$ has $k(N-k)$ cross pairs, and a pair is \emph{violated} if the compression reverses its order so that $r^z_i(q)<r^z_i(p)$. Let $p_k(i)$ be the fraction of violated cross pairs of $i$, and let $\bar p_k=\frac1n\sum_i p_k(i)$.

The theorem below states that a memory cannot lose many neighbors at scale $k$ unless it reverses many cross pairs.

\begin{theorem}[Rank-violation bound]\label{thm:bound}
For every compression and every $1\le k\le N-1$,
\begin{equation}\label{eq:bound}
1-\nu_k\;\le\;\frac{N}{\sqrt{k(N-k)}}\cdot\frac1n\sum_{i=1}^n\sqrt{p_k(i)}\;\le\;\frac{N}{\sqrt{k(N-k)}}\,\sqrt{\bar p_k}.
\end{equation}
\end{theorem}
\begin{proof}
Fix $i$ and let $m_i=k-|\mathcal N^x_k(i)\cap\mathcal N^z_k(i)|$. Exactly $m_i$ items are \emph{lost}, in $\mathcal N^x_k(i)$ and outside $\mathcal N^z_k(i)$, and exactly $m_i$ are \emph{gained}, in $\mathcal N^z_k(i)$ and outside $\mathcal N^x_k(i)$, because both sets have size $k$. Take a lost $p$ and a gained $q$. Then $p$ is near and $q$ is far in the uncompressed memory, so $(p,q)$ is a cross pair, and $r^z_i(q)\le k<r^z_i(p)$, so it is violated. These $m_i^2$ pairs are distinct, hence $m_i^2\le p_k(i)\,k(N-k)$ and $m_i\le\sqrt{p_k(i)\,k(N-k)}$. By~\eqref{eq:dlp}, $1-\DLP_k=\frac{1}{nk}\sum_im_i$, and by~\eqref{eq:nu}, $1-\nu_k=\frac{N}{N-k}(1-\DLP_k)$. Combining,
\[
1-\nu_k=\frac{N}{nk(N-k)}\sum_{i=1}^n m_i\le\frac{N}{\sqrt{k(N-k)}}\cdot\frac{1}{n}\sum_{i=1}^n\sqrt{p_k(i)}.
\]
The second inequality is Jensen's inequality for the concave square root.
\end{proof}

The middle term of~\eqref{eq:bound} is $\frac{N}{\sqrt{k(N-k)}}\,\overline{\sqrt{p_k}}$, with $\overline{\sqrt{p_k}}=\frac1n\sum_i\sqrt{p_k(i)}$. This bound is strongest where $G$ is read. Because the factor $N/\sqrt{k(N-k)}$ equals $1/\sqrt{(k/N)(1-k/N)}$, it lies between 2 and 2.18 across the $G$ band ($0.3\le k/n\le0.5$). Consequently, a small violation rate at these scales forces $\nu_k$ close to 1, transferring the bound directly to $G$. By contrast, at $k=10$ the factor is approximately $\sqrt{N/10}$, which reaches 14 for a memory of 2,000 items. Theorem~\ref{thm:bound} thus singles out the middle and far scales as the scales where ordering cross pairs controls the profile, motivating LAQ to train its codes at those scales. On the held-out memories of Section~\ref{sec:laqresults}, the bound holds across all 160 combinations of memory, quantizer, and scale (Figure~\ref{fig:samelaq}e).

Because rank violations are counted by a step function, we bound them with a hinge on distance gaps. Let $\Delta^x=\|x_i-x_q\|^2-\|x_i-x_p\|^2\ge0$ be the uncompressed gap of a cross pair, $\Delta^z=\|z_i-z_q\|^2-\|z_i-z_p\|^2$ be its compressed gap, and $h=[\eta\Delta^x-\Delta^z]_+$ with margin $\eta>0$. Pushing compressed gaps above a fraction of the uncompressed gaps drives the violation rate down, as the following corollary shows.

\begin{corollary}[Hinge surrogate]\label{cor:hinge}
Let $(p,q)$ be a uniformly random cross pair of a uniformly random item at scale $k$, and $\bar H_k=\mathbb E\,h$. For every $\delta>0$,
\begin{equation}\label{eq:hinge}
\bar p_k\;\le\;\Pr(\Delta^x<\delta)+\frac{\bar H_k}{\eta\,\delta}.
\end{equation}
\end{corollary}

The first term depends on the uncompressed memory alone, measuring the share of cross pairs whose two squared distances to $i$ lie within $\delta$ of each other. The second term is the expected hinge, for which the average hinge over uniformly drawn cross pairs provides an unbiased estimate. Together, Theorem~\ref{thm:bound} and Corollary~\ref{cor:hinge} turn the profile at a chosen scale into an objective that gradient descent can lower.

\subsection{Predicting the far-end loss}\label{sec:noise}

Theorem~\ref{thm:bound} says that the far end is kept when few cross pairs are reversed. The next result says how many are reversed. For anchor $i$, let $D_j=\|x_i-x_j\|^2$ and $\tilde D_j=\|z_i-z_j\|^2$ for $j\ne i$, and let $m_i$ and $s_i$ be the mean and standard deviation of the $D_j$. Fit the compressed distances by the least-squares line $\tilde D_j\approx a_iD_j+b_i$ with $a_i>0$. A common map of this form leaves every rank unchanged, so a pair can be reversed by the residual $e_j=(\tilde D_j-b_i)/a_i-D_j$ alone. We call $\omega_i=\mathrm{sd}(e_j)/s_i$ the \emph{relative distance noise} of anchor $i$: the distortion of its distances measured against their spread. Let $\phi$ and $\Phi$ be the density and distribution function of $\mathcal N(0,1)$, and $z_\alpha=\Phi^{-1}(\alpha)$.

\begin{proposition}[Noisy order]\label{prop:noise}
Fix $\alpha=k/n\in(0,1)$. Suppose the standardized distances $u_j=(D_j-m_i)/s_i$ of anchor $i$ are independent $\mathcal N(0,1)$, and the scaled residuals $e_j/s_i$ are independent normal with mean 0 and variance $\omega_i^2$, independent of the $u_j$. Then, as $n\to\infty$, the expected violation rate of anchor $i$ at scale $k$ converges to
\begin{equation}\label{eq:noise}
\pi(\alpha,\omega_i)=\frac{1}{\alpha(1-\alpha)}\int_{-\infty}^{z_\alpha}\!\int_{z_\alpha}^{\infty}\phi(u)\,\phi(v)\,\Phi\!\Bigl(-\frac{u-v}{\sqrt2\,\omega_i}\Bigr)\,du\,dv,
\end{equation}
and $\pi(\alpha,\omega)=\phi(z_\alpha)^2\,\omega^2/\bigl(2\alpha(1-\alpha)\bigr)+O(\omega^3)$ as $\omega\to0$.
\end{proposition}

Proposition~\ref{prop:noise} counts reversed pairs. The same model gives the profile itself.

\begin{corollary}[Profile under noisy order]\label{cor:noiseprofile}
Under the assumptions of Proposition~\ref{prop:noise}, as $n\to\infty$ the contribution of anchor $i$ to $\nu_k$ converges to
\begin{equation}\label{eq:noiseprofile}
\nu(\alpha,\omega_i)=\frac{\Phi_2(z_\alpha,z_\alpha;\varrho_i)-\alpha^2}{\alpha(1-\alpha)},\qquad \varrho_i=\bigl(1+\omega_i^2\bigr)^{-1/2},
\end{equation}
where $\Phi_2(\cdot,\cdot;\varrho)$ is the bivariate normal distribution function with unit variances and correlation $\varrho$. It decreases strictly in $\omega_i$, and at $\alpha=1/2$ it equals $\frac2\pi\arcsin\varrho_i$, as in Proposition~\ref{prop:sheppard}.
\end{corollary}

The far end thus depends on one ratio: how much the compression perturbs the distances from an anchor, against how much those distances spread. At the same distortion, a memory whose distances spread less loses more of its far end, and any quantizer that lowers $\omega_i$ in the $G$ band raises $\nu_k$ there. The model rests on a normal approximation of each anchor's distances. In the $G$ band, these distances are sums over many directions and lie in the bulk of their distribution. At $k=10$, they lie in the lower tail, where the local structure of the memory decides. The independence of the residuals also holds approximately: items that share a codeword have correlated errors, and the noise varies across items. Figure~\ref{fig:noise} tests the consequences of both approximations.

The held-out memories bear out both results (Figure~\ref{fig:noise} in Appendix~\ref{app:measure}). Standardized by their normal mean and variance, the distances of each anchor match the normal quantile to within a median of 0.10 standard deviations at $k=0.4n$ and 0.06 at $k=0.5n$, against 0.42 at $k=10$. Across all 52 held-out runs, for the OPQ start and LAQ, $\pi(0.4,\omega_i)$ predicts the measured violation rate at $k=0.4n$ with correlation 0.979. The median error is 5.3 percent with the retrieval-tuned encoders, and the prediction is 25 to 31 percent low with Titan V2. These predictions use the full integral~\eqref{eq:noise}. Its leading term is larger, by a median of 25 percent with Titan V2, so the gap with Titan V2 lies in the assumptions of the model and not in the expansion. At $k=10$, the normal model predicts 2.8 times the measured rate (median), so the near end keeps more order than the model allows. Evaluated at the mean relative distance noise of each run, Corollary~\ref{cor:noiseprofile} gives $\nu_k$ at $k=0.4n$ within a median of 0.005 of the measured value (correlation 0.982). The relative distance noise ranges from 0.16 to 0.37 with the retrieval-tuned encoders and from 0.47 to 0.52 with Titan V2, whose OPQ start keeps the least of the far end. The data entropy of a memory~\citep{han2024hft}, $h=-\sum_\ell u_\ell\log_2u_\ell$ with $u_\ell$ the singular values of the standardized memory divided by their sum, tells how much LAQ lowers this noise. Within each retrieval-tuned encoder, across its 16 held-out runs, $h$ has a rank correlation of 0.77 to 0.81 with the drop in mean relative distance noise from OPQ to LAQ and of 0.82 to 0.84 with the gain in $G$ ($p\le4\times10^{-4}$ in each case). Memories of higher data entropy spread their information over more directions, and LAQ removes more of their distance noise.

%% file: t_laq.tex
\section{Locality-aware quantization}\label{sec:laq}

Locality-aware quantization (LAQ) trains the codebooks of an OPQ index to keep the order of cross pairs at the middle and far scales, the target identified by Theorem~\ref{thm:bound}. Whereas product quantizers learn codebooks to reconstruct vectors or keep nearest neighbors, LAQ adds this far-end target, needs no labels, and keeps the rest of the OPQ index intact. Figure~\ref{fig:memgraph}g shows the procedure.

\paragraph{Starting point} LAQ first fits PCA on the training memory in the full basis, followed by OPQ with $M=d/8$ subspaces of eight dimensions and $K=256$ codewords each, storing every item in $d/8$ bytes. Let $\mu$ be the mean of the training vectors and $R$ be the orthogonal map combining the PCA basis with the OPQ rotation. Let $y_i=R(x_i-\mu)$ denote the rotated item, $y_{i,\ell}$ its $\ell$-th subvector, and $C=\{c_{\ell,j}\}$ ($\ell=1,\dots,M$, $j=1,\dots,K$) the codebooks initialized at the OPQ values $C_0$. Each item is encoded by its nearest codeword in each subspace and decoded as the concatenation $z_i(C)$. Because $R$ is orthogonal, distances in the rotated basis equal those in the original space.

\paragraph{Objective} LAQ minimizes
\begin{equation}\label{eq:J}
J(C)=\frac{\mathcal L_{\mathrm{rec}}(C)}{\mathcal L_{\mathrm{rec}}(C_0)}+\lambda_{\mathrm{mid}}\,\mathcal L_{\mathrm{mid}}(C)+\lambda_{\mathrm{far}}\,\mathcal L_{\mathrm{far}}(C),
\end{equation}
where $\mathcal L_{\mathrm{rec}}=\frac1n\sum_i\|y_i-z_i\|^2$ is the reconstruction error, and $\mathcal L_{\mathrm{mid}}$ and $\mathcal L_{\mathrm{far}}$ are hinge losses evaluated at two scale groups with weights $\lambda_{\mathrm{mid}}$ and $\lambda_{\mathrm{far}}$. For a scale group $\mathcal S\in\{\mathrm{mid},\mathrm{far}\}$ and a batch of sampled cross pairs $(p,q)$ with anchors $i$,
\begin{equation}\label{eq:hingeloss}
\mathcal L_{\mathcal S}=\frac{1}{s_{\mathcal S}}\,\mathrm{mean}\bigl[\eta\Delta^x-\Delta^z\bigr]_+,
\end{equation}
where $\Delta^x$ and $\Delta^z$ are the distance gaps defined in Corollary~\ref{cor:hinge}. The normalizer $s_{\mathcal S}$ is the median uncompressed gap of the group, computed once at initialization so that both groups contribute on a common scale. The middle group draws $k/n\in\{1\%,5\%\}$, while the far group draws $k$ uniformly from $\mathcal K$, the $G$ band. The normalized reconstruction term keeps the near end, which product quantization already serves well.

\paragraph{Sampling cross pairs} Each step draws 1,024 anchors per group, and for each anchor samples a scale $k$ and a cross pair. Half of the pairs are drawn uniformly: $p$ from the $k$ nearest items of $i$, and $q$ from the remaining $N-k$ items. By Corollary~\ref{cor:hinge}, these pairs yield an unbiased estimate of $\bar H_k$, which bounds the violation rate and, through Theorem~\ref{thm:bound}, bounds $1-\nu_k$. The other half straddle rank $k$: $p$ is drawn from ranks $k-w+1$ to $k$ and $q$ from ranks $k+1$ to $k+w$, with $w=\max(3,\lfloor0.1k\rfloor)$. Because lost and gained neighbors arise at this boundary, as shown in the proof of Theorem~\ref{thm:bound}, these boundary pairs focus half of the gradient where the neighborhood changes.

\paragraph{Straight-through codes} Because encoding is a hard assignment, LAQ uses hard assignments in the forward pass and the soft assignment $\mathrm{softmax}(-\|y_{i,\ell}-c_{\ell,j}\|^2/T)$ in the backward pass~\citep{chen2020dpq}. It re-encodes all training items with the current codebooks at every step. The temperature at step $t$ decays as $T_t=T_0\cdot0.05^{t/t_{\max}}$ over $t_{\max}$ steps, where $T_0$ is the initial mean squared distance from a subvector to its nearest codeword. The codebooks are updated using Adam~\citep{kingma2015adam} with a cosine-decayed learning rate equal to $0.002$ times the standard deviation of the entries of $C_0$.

\paragraph{Checkpoint on a monitor set} LAQ draws 2,000 training items as a monitor set. Every 250 steps, it evaluates on this monitor set both $\nu_k$ at $k/n=0.4$ on the set (the midpoint of the $G$ band) and the relative reconstruction error $\mathcal L_{\mathrm{rec}}(C)/\mathcal L_{\mathrm{rec}}(C_0)$. It keeps the codebooks with the highest $\nu_k$ among checkpoints whose relative reconstruction error is at most 1.15, falling back to $C_0$ if no checkpoint improves upon the start.

\paragraph{The far hinge at the OPQ start} At the OPQ start, the far hinge can fall while the reconstruction error holds to first order. Let $n_{\ell j}$ be the number of training items whose $\ell$-th subvector is assigned to $c_{\ell,j}$, and let $\langle\cdot,\cdot\rangle$ be the inner product of codebooks.

\begin{proposition}[Far hinge at no first-order cost]\label{prop:start}
Suppose every codeword of $C_0$ is the mean of the training subvectors assigned to it, no subvector is equidistant from its two nearest codewords, and $\mathcal L_{\mathrm{far}}$, computed with the assignments of $C_0$, has a gradient $g$ at $C_0$. Then, for every direction $h$ with $\langle g,h\rangle>0$ and all small enough $\epsilon>0$, the step $C_\epsilon=C_0-\epsilon h$ satisfies
\begin{equation}\label{eq:start}
\begin{aligned}
\mathcal L_{\mathrm{rec}}(C_\epsilon)&=\mathcal L_{\mathrm{rec}}(C_0)+\epsilon^2\sum_{\ell,j}\frac{n_{\ell j}}{n}\,\|h_{\ell j}\|^2,\\
\mathcal L_{\mathrm{far}}(C_\epsilon)&=\mathcal L_{\mathrm{far}}(C_0)-\epsilon\,\langle g,h\rangle+o(\epsilon).
\end{aligned}
\end{equation}
\end{proposition}

Proposition~\ref{prop:start} is a statement about the two training terms: along any direction that lowers the far hinge, the reconstruction error rises at second order in the step. This holds for $h=g$ and for the first step of Adam, which is $h=\mathrm{sign}(g)$ up to scale, with $\langle g,\mathrm{sign}(g)\rangle=\|g\|_1>0$.

\paragraph{Straight-through steps} Training computes a different gradient. The straight-through codes use the hard assignments of Proposition~\ref{prop:start} in the forward pass and the soft assignments at temperature $T$ in the backward pass, so the computed gradient $g_{\mathrm{st}}$ differs from $g$. As $T\to0$ the soft assignments become hard and $g_{\mathrm{st}}\to g$. At the starting temperature $T_0$, we measured both gradients on the same 8,192 far cross pairs at all 52 OPQ starts of Section~\ref{sec:laqresults}. The cosine between $g_{\mathrm{st}}$ and $g$ is 0.56 to 0.69 (median 0.62), so $h=g_{\mathrm{st}}$ meets the condition of Proposition~\ref{prop:start} at every start. One first Adam step along $g_{\mathrm{st}}$, re-encoding all items, lowers the far hinge by 1.1 to 2.5 percent (median 1.6 percent) and raises the reconstruction error by at most 0.008 percent, while 0.05 to 0.11 percent of the subvectors change codeword.

\paragraph{From the hinge to $G$} A lower far hinge lowers the bound of Corollary~\ref{cor:hinge} on the violation rate in the $G$ band, and through Theorem~\ref{thm:bound} the bound on $1-\nu_k$. The checkpoint rule keeps the gain on the monitor set: the start is itself a candidate, so the kept codebooks have a monitor $\nu_k$ at least that of $C_0$, at a reconstruction error at most 1.15 times that of $C_0$. Section~\ref{sec:laqresults} tests the gain on held-out memories. The OPQ starts are close to the fixed point that Proposition~\ref{prop:start} assumes. On all 52 training memories, one more centroid step lowers the reconstruction error by at most 0.032 percent (median 0.016 percent) and moves the codebooks by at most 0.6 percent of their norm. Corollary~\ref{cor:noiseprofile} also predicts the size of the gain. Evaluated at the mean relative distance noise of OPQ and of LAQ, it predicts a mean gain in $\nu_k$ at $k=0.4n$ of 0.0126, against 0.0124 measured on the held-out memories, with a rank correlation of 0.92 across runs and a predicted gain above zero in all 52.

\paragraph{What LAQ changes} LAQ modifies the codebook values and nothing else: the PCA basis, rotation, bytes per item, and asymmetric distance computation (ADC) search all remain identical to those of OPQ. An LAQ index therefore replaces an OPQ index directly and adds no operation at query time. A single recipe serves every memory and encoder: $t_{\max}=1{,}000$ steps, $\lambda_{\mathrm{mid}}=\lambda_{\mathrm{far}}=1$, and $\eta=1$ (Appendix~\ref{app:recipe}).

\paragraph{Complexity} Let $n$ be the number of training items, $d$ the dimension, $K=256$ the codebook size, $B=2{,}048$ the number of cross pairs per step (1,024 anchors in each of the two groups), and $t_{\max}$ the number of steps. Ranking the training memory once costs $O(n^2(d+\log n))$ time and $O(n^2)$ memory. Each step encodes all items with the current codebooks in $O(nKd)$ time and evaluates the hinge on $B$ cross pairs in $O(Bd)$ time. Training therefore costs $O(n^2(d+\log n)+t_{\max}(nKd+Bd))$ time in total, while the query cost and the encoding of a new item, $O(Kd)$, remain those of OPQ. The $n\times n$ matrix of 7,336 to 7,672 items fits on one 48~GB GPU, and the 1,000 steps take 35 to 112 seconds on Multi-Doc-2025. For a memory of a million items, a scalable variant ranks against $m=8{,}192$ reference items at each step and decodes only the items used in that step, with a reconstruction batch of $b=16{,}384$. This reduces the per-step cost to $O(Bm(d+\log m)+(B+b)Kd)$ and memory to $O(nd+Bm)$ (Section~\ref{sec:marco}).

%% file: t_setup.tex
\section{Experimental setup}\label{sec:setup}
We evaluate the two fidelities across five memories and six encoders, and test LAQ on held-out splits of four memories as well as on a million-passage memory.

\paragraph{Memories} Table~\ref{tab:corpora} lists five memories: three agent memories and two topic corpora. FinQA, LoCoMo, and the topic corpora serve the measurement, and Multi-Doc-2025 and two additional memories test LAQ. \emph{FinQA}~\citep{chen2021finqa} contains pages from the annual reports of S\&P 500 companies paired with numerical questions annotated with supporting facts. We retain the 106 companies in the current index and up to three report pages per company. Every sentence and table row is one item, prefixed by the company name and fiscal year. This yields 8,356 items across 11 Global Industry Classification Standard (GICS) sectors and 863 questions. \emph{Multi-Doc-2025}~\citep{multidoc2025} is a financial question-answering benchmark spanning 179 10-K filings from 89 S\&P 500 companies for fiscal years 2022 to 2024. Each paragraph and table row serves as a candidate item with the same prefix. Every filing retains the passages containing its questions' evidence plus a random sample, up to 56 passages, yielding 10,024 items and 898 questions with located evidence. Multi-Doc-2025 is used to test LAQ. \emph{LoCoMo}~\citep{maharana2024locomo} is a long-term conversational memory benchmark for agents, comprising ten conversations of 19 to 32 sessions, 5,882 turns, and 1,536 answerable questions annotated with their evidence turns. Each turn, prefixed by its speaker, forms one item. \emph{20 Newsgroups}~\citep{lang1995newsweeder} and \emph{AG News}~\citep{zhang2015charcnn} contribute 10,000 random documents each, spanning 20 and 4 topics, respectively. A sixth memory, \emph{FinDER}~\citep{choi2025finder}, is reserved for testing LAQ: it pairs 5,703 queries written by financial professionals with expert-selected evidence from the 10-K filings of 497 S\&P 500 companies. Following our Multi-Doc-2025 construction, we keep up to 30 passages per filing. Because FinDER provides the evidence text without specifying its company, we locate each excerpt in the filing that contains the most of its rarer words. When multiple filings match, the company named in the query breaks the tie, and any excerpt located in a company other than the one named is dropped. Taking the one or two passages in that filing whose words appear in the excerpt as the evidence passages yields 14,949 items and 5,129 questions. Finally, a seventh memory comprising 1,000,000 passages from MS MARCO~\citep{bajaj2016msmarco} tests LAQ at scale (Section~\ref{sec:marco}).

\begin{table}[t]
\centering\footnotesize
\caption{Five memories with company, session, or topic structure, three with annotated questions.}\label{tab:corpora}
\setlength{\tabcolsep}{3pt}
\begin{tabular}{@{}lrrl@{}}
\toprule
Memory & Items & Questions & Structure used \\
\midrule
FinQA & 8,356 & 863 & company, sector, fiscal year \\
Multi-Doc-2025 & 10,024 & 898 & company, sector, fiscal year \\
LoCoMo & 5,882 & 1,536 & conversation, session \\
20 Newsgroups & 10,000 & -- & topic, topic family \\
AG News & 10,000 & -- & topic \\
\bottomrule
\end{tabular}
\end{table}

\paragraph{Encoders} Six encoders serve the fidelity measurement runs. Three are retrieval-tuned sentence encoders used with their query instructions: BGE-base-en-v1.5 (BGE-base) and BGE-large-en-v1.5 (BGE-large)~\citep{xiao2024cpack} at 768 and 1,024 dimensions, and E5-base-v2 (E5-base)~\citep{wang2022e5} at 768 dimensions. Three are encoders of other kinds: the text encoder of MobileCLIP2-L/14~\citep{faghri2025mobileclip2} (768 dimensions), RoBERTa-base~\citep{liu2019roberta} with mean pooling (768), and averaged GloVe vectors~\citep{pennington2014glove} (300). A seventh encoder, the commercial Amazon Titan Text Embeddings V2 (Titan V2)~\citep{titanbinary2024} at 1,024 dimensions, is used on Multi-Doc-2025 to test LAQ. All vectors have unit length.

\paragraph{Compressions} There are 18 measurement runs: the six encoders on the two agent memories FinQA and LoCoMo, and the three encoders of other kinds on the two topic corpora. Each run applies 27 compressions fitted on the memory itself. Scalar codes comprise half precision along with 8- and 4-bit codes. Binary codes comprise 1-bit sign codes, the same codes with full-precision rescoring of the 40 nearest candidates, and 2- and 4-bit thermometer codes that set one bit per threshold passed on each coordinate. Truncation, PCA, and Gaussian random projection reduce vectors to $d/2$, $d/4$, $d/8$, and $d/16$ dimensions. In addition, PCA to $d/4$ and $d/8$ is stored with 8-bit codes, and PCA to $d$ and $d/2$ is followed by a random rotation and binarization. Product quantization uses $d/8$, $d/16$, and $d/32$ bytes, while UMAP~\citep{mcinnes2018} maps vectors to 16 dimensions. Finally, OPQ with and without rescoring is included at $d/8$ bytes. For comparison, t-SNE~\citep{vandermaaten2008}, SR-tSNE~\citep{han2026srtsne}, PHATE~\citep{moon2019}, Isomap~\citep{tenenbaum2000}, and nonnegative PCA (NPCA)~\citep{han2010tcbb} are fitted on FinQA and LoCoMo with BGE-base (Table~\ref{tab:nonlinear}). Binary codes are compared by Hamming distance and all other codes by Euclidean distance on the decoded vectors. In our figures, PCA with 8-bit codes is labeled PCA + int8 and rescored binary codes binary + rescoring. Binarization after PCA and a random rotation is labeled PCA + rotation + binary (PCA + rot. + binary where space is short).

\paragraph{Operations} Each operation is executed on both the compressed and uncompressed memories, and we report their agreement. We model an agent's memory operations directly on the vectors, which scores every operation on every compression without a language model in the loop.
\begin{itemize}[leftmargin=1.2em,itemsep=1pt]
\item \emph{Retrieval.} Each question retrieves its 10 nearest items (within its conversation for LoCoMo). Agreement is the overlap of the two top-10 lists, while quality is evidence recall@10, the share of a question's evidence among those 10 items.
\item \emph{Consolidation into themes.} We run $k$-means on the compressed memory, initialized from the centroids of the uncompressed partition, using 11 themes for the financial memories and 20 and 4 themes for the topic corpora. Agreement is measured by the adjusted Rand index (ARI)~\citep{hubert1985}, which we call theme agreement.
\item \emph{Deletion.} For each session or topic, the memory selects as many items as the session or topic contains, choosing those nearest its centroid. This models how a memory identifies deletion candidates by meaning when metadata are incomplete. Agreement is the Jaccard overlap of the deleted sets.
\item \emph{Company operations.} The information barrier walls off a company by applying deletion to it, while peer groups consist of each company's five nearest companies by centroid. The company map is the rank correlation between distances among compressed company centroids and the uncompressed ones; the sector, session, and topic maps are defined in the same way.
\item \emph{Global references.} These use the subsamples of $G$. \emph{All distances} is the Spearman correlation across 200,000 pairwise distances~\citep{kobak2019}. \emph{Triplets} is the accuracy on random triplets~\citep{wang2021}. The \emph{median split} is $\nu_{k^\star}$ from Proposition~\ref{prop:duality}.
\end{itemize}

Because consolidation, deletion, the session, sector, and topic maps, and the global references compare each item against the whole memory, we call them the far-end operations. Across the 27 compressions of each run, we relate each operation's agreement to $L$ and $G$ using Spearman correlation as well as the partial rank correlation with one fidelity given the other.

\paragraph{Held-out protocol for LAQ} LAQ is trained and tested on disjoint companies or conversations. For FinQA and Multi-Doc-2025, each split holds out a quarter of the companies (26 of 106 and 22 of 89). For LoCoMo it holds out two of the ten conversations. Three splits (seeds 2026 to 2028) were fixed in advance. A fourth split (seed 2029) was drawn after the LAQ recipe was fixed and run once with the recipe unchanged. PCA, OPQ, and LAQ are fitted on the training part only, and every reported number is computed on the held-out part using its own questions. Evaluating the three retrieval-tuned encoders across all three memories and Titan V2 on Multi-Doc-2025 yields 40 held-out runs, each comparing LAQ against its own OPQ start at $d/8$ bytes. The other quantizers were evaluated on the first three splits. On Multi-Doc-2025, queries remain in full precision and are searched via ADC, just as an OPQ index serves them. $L$, $G$, and the far-end operations are evaluated on the stored codes. FinDER was run after all of the above, with the recipe unchanged. Each of its four splits (seeds 2026 to 2029) holds out 124 of its 497 companies, with queries searched by ADC. The three retrieval-tuned encoders thus provide 12 further runs, which we report separately from the 40.

\paragraph{Quantizers compared} On the held-out Multi-Doc-2025 memories, we compare LAQ at $d/8$ bytes against OPQ and eight other quantizers. Five store codes of approximately the same size: PQ~\citep{jegou2011pq}, RQ~\citep{chen2010rvq}, LSQ++~\citep{martinez2018lsq}, PCA to $d/8$ dimensions with 8-bit codes, and RaBitQ~\citep{gao2024rabitq} with $d$ bits plus its per-vector factors (104 and 136 bytes). Sign and ITQ~\citep{gong2013itq} binary codes are rescored on the 40 nearest candidates, and ScaNN uses anisotropic quantization~\citep{guo2020scann}. All but ScaNN are fitted on the training companies.

%% file: t_results_measure.tex
\section{What the two ends measure}\label{sec:measure}

Each end of the profile predicts its own operations: retrieval follows $L$, consolidation and deletion follow $G$, and compressions of equal size can keep very different far ends. Section~\ref{sec:laqresults} tests whether LAQ raises the far end at the same bytes.

\subsection{Compressions separate on the $(L,G)$ plane}

\begin{figure*}[p]
\centering
\includegraphics[width=\textwidth,height=0.8\textheight,keepaspectratio]{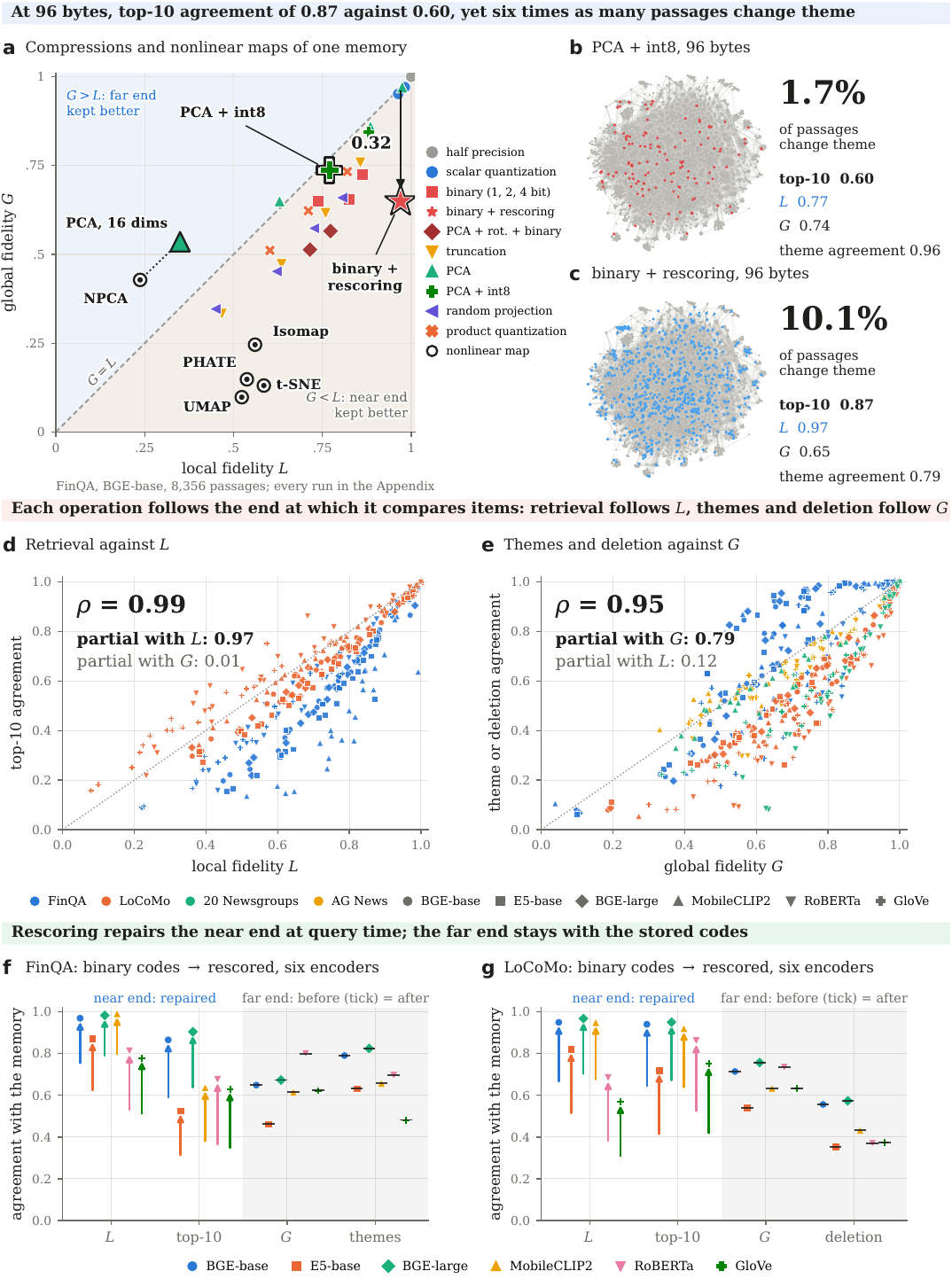}
\caption{At the same bytes, a compression that keeps more nearest neighbors can displace more of the overall memory, because retrieval follows $L$ whereas themes and deletion follow $G$ (local and global fidelity). (a) 27 compressions of FinQA (BGE-base) alongside four nonlinear maps (rings: Isomap~\citep{tenenbaum2000}, NPCA~\citep{han2010tcbb}, PHATE~\citep{moon2019}, t-SNE); triangle: PCA to 16 dimensions. Rescored binary codes lie 0.32 below the diagonal $G=L$, while t-SNE, PHATE, and UMAP keep $G$ at 0.22 or less. (b, c) Two 96-byte compressions plotted on the uncompressed memory layout; colored points mark passages moved to another theme: 1.7 percent for PCA with 8-bit codes versus 10.1 percent for rescored binary codes. (d, e) Across every compression and run, retrieval follows $L$ (median Spearman $\rho$ 0.99; partial 0.97 given $G$), whereas themes and deletion follow $G$ ($\rho$ 0.95; partial 0.79 given $L$). (f, g) Rescoring the 40 nearest binary codes raises $L$ and top-10 agreement while leaving $G$, themes, and deletion unchanged at the binary values (black ticks). All runs: Appendix~\ref{app:measure}.}\label{fig:measure}
\end{figure*}

Compressions of equal size keep different ends of a memory. Figure~\ref{fig:measure}a places every compression of FinQA with BGE-base on the $(L,G)$ plane; in Appendix~\ref{app:measure}, Figure~\ref{fig:plane} shows both agent memories with the retrieval-tuned encoders, and Figure~\ref{fig:otherenc} shows the other three encoders. Within a given memory, the two ends correlate across compressions (Spearman 0.81 to 0.95 over the 18 runs) because fewer bytes lower both. Yet compressions with the same $L$ can differ widely in $G$. With the retrieval-tuned encoders, half precision and 8-bit scalar codes keep both ends at 0.97 or above. PCA, PCA with 8-bit codes, and product quantization keep $G$ within 0.12 of $L$, whereas binarization after PCA and a random rotation places $G$ 0.06 to 0.22 below $L$. Among the linear compressions, rescored binary codes fall furthest below the diagonal, by 0.21 to 0.41. At 96 bytes per passage on FinQA with BGE-base, they keep $L$ at 0.97 but move 10.1 percent of the passages to another theme. PCA with 8-bit codes keeps $L$ at 0.77 and moves only 1.7 percent (Figure~\ref{fig:measure}b, c). Among nonlinear maps with BGE-base, UMAP to 16 dimensions keeps $L$ at 0.36 to 0.54 and $G$ at 0.10 to 0.20, while t-SNE and PHATE keep $G$ at 0.22 or less. Isomap keeps $G$ at 0.25 and 0.43. NPCA lies above the diagonal, with $G$ at 0.43 and 0.44 and $L$ at 0.22 and 0.24, as does PCA to the same 16 dimensions, with $G$ at 0.54 and $L$ at 0.35 (Figure~\ref{fig:measure}a, Table~\ref{tab:nonlinear}). With the encoders of other kinds, PCA and product quantization lie on or above the diagonal on both the agent memories and the topic corpora (Figures~\ref{fig:otherenc} and~\ref{fig:topics}a--f). One-bit codes of GloVe and RoBERTa vectors reach $G$ of 0.58 to 0.81 at $L$ of 0.26 to 0.51. Because these encoders have coordinates of highly unequal variance, taking the sign of a low-variance coordinate perturbs the nearest neighbors far more than the nearer half.

\subsection{Each end predicts its own operations}

Retrieval follows $L$, whereas the far-end operations follow $G$ (Figure~\ref{fig:measure}d, e); Figure~\ref{fig:predict} in Appendix~\ref{app:measure} plots each operation against both fidelities. Top-10 agreement lies close to a single increasing curve in $L$, with a median within-run Spearman correlation of 0.99 versus 0.89 for $G$. Conversely, the far-end operations lie closer to a single curve in $G$, with a median correlation of 0.95 versus 0.87 for $L$. Because $L$ and $G$ fall together under more severe compression, partial correlations separate the two ends even more sharply.

\paragraph{Partial correlations} This separation holds when conditioning on the other end: retrieval follows $L$ given $G$, and with the retrieval-tuned encoders, consolidation and deletion follow $G$ given $L$. Figure~\ref{fig:partial} in Appendix~\ref{app:measure} gathers every operation across every run. Retrieval sits at the lower right, with a partial correlation with $L$ given $G$ of 0.83 to 0.99 (Tables~\ref{tab:pred} and~\ref{tab:predb} in Appendix~\ref{app:measure}). Consolidation, deletion, and the session and topic maps sit at the upper left. With the retrieval-tuned encoders, consolidating the financial memory into themes follows $G$. Its partial correlations with $G$ given $L$ are 0.68, 0.76, and 0.66 for BGE-base, E5-base, and BGE-large, whereas with $L$ given $G$ they are 0.04, 0.01, and $-0.02$. Deleting a LoCoMo session has partial correlations of 0.86, 0.55, and 0.93 with $G$ given $L$, versus $-0.11$, 0.36, and $-0.14$ with $L$ given $G$. The session and topic maps follow $G$ at 0.56 to 0.93 given $L$, and the two global references, all distances and triplets, follow $G$ at 0.78 to 1.00. The operations defined on companies sit between the two ends. Peer groups, the barrier, and the company and sector maps compare 106 company centroids, making each a nearest-neighbor question in the space of companies. With the retrieval-tuned encoders, they follow $L$ (0.39 to 0.65 given $G$) at least as closely as $G$ (0.21 to 0.51 given $L$). In short, an operation follows the end of the profile at which it compares items.

\subsection{Rescoring repairs the near end and keeps the stored codes}

Rescoring repairs the near end while leaving the far end tied to the stored codes. Disk-based vector search keeps binary codes in memory and rescores candidates using full-precision vectors read from disk; Figure~\ref{fig:measure}f, g reproduces this pipeline using the 40 nearest codes. On FinQA, rescoring raises $L$ from 0.74 to 0.97 with BGE-base, from 0.61 to 0.87 with E5-base, and from 0.77 to 0.98 with BGE-large. It lifts top-10 agreement from 0.57 to 0.87, 0.30 to 0.53, and 0.62 to 0.90. Yet $G$ remains fixed at the value of the binary codes, as does every operation computed on them. Theme agreement on the financial memory stays at 0.79, 0.63, and 0.82, and LoCoMo session deletion stays at 0.56, 0.35, and 0.57. Judged by a recall check, rescored binary codes appear close to the uncompressed memory; judged by $G$, they keep only the far end of plain binary codes. The same pattern holds for every code family that preserves Euclidean distances (Table~\ref{tab:rescoreall}).

\paragraph{Evidence and fiscal year} With the retrieval-tuned encoders, evidence recall follows $L$, with partial correlations of 0.71 to 0.83 given $G$ on FinQA and 0.61 to 0.87 on LoCoMo (Figure~\ref{fig:rescore}c in Appendix~\ref{app:measure}). Because financial text repeats across years, a retrieved passage can come from the right company but the wrong fiscal year. In the uncompressed memories with the retrieval-tuned encoders, 0.18 of the right-company passages come from another fiscal year, and every linear compression keeps this share between 0.15 and 0.24 (Figure~\ref{fig:rescore}d).

\subsection{Same bytes, different ends}\label{sec:same}

\begin{figure*}[p]
\centering
\includegraphics[width=\textwidth,height=0.8\textheight,keepaspectratio]{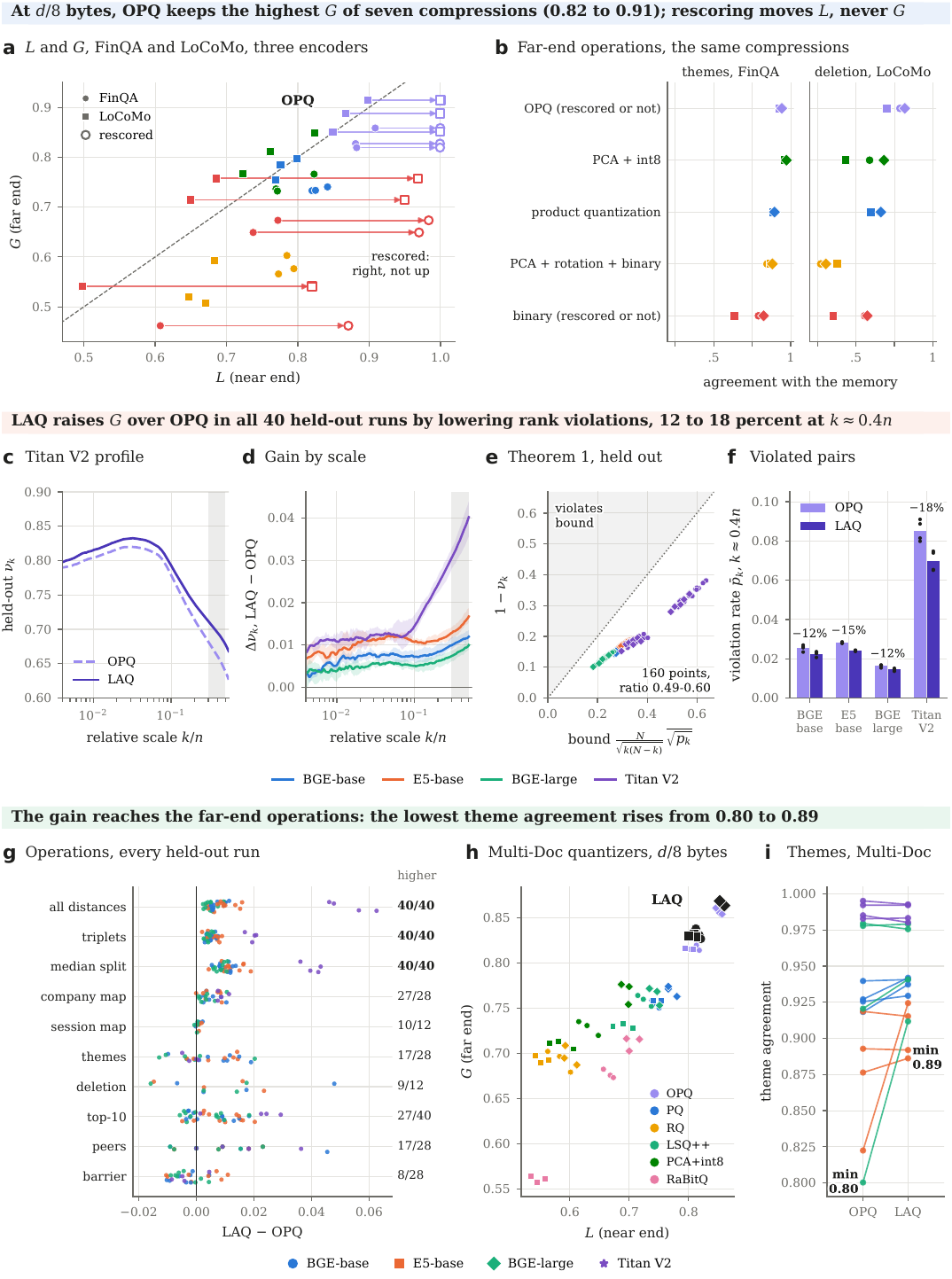}
\caption{At the same bytes, OPQ keeps the far end best, and LAQ raises it further on every held-out memory by lowering rank violations ($L$, $G$: local and global fidelity). (a, b) Seven compressions at $d/8$ bytes on FinQA and LoCoMo with the retrieval-tuned encoders: OPQ has the highest $G$ (0.82 to 0.91), whereas rescoring (arrows to open markers) raises $L$ and top-10 agreement but leaves $G$ and the far-end operations unchanged. (c--f) Held-out Multi-Doc-2025 across four splits: LAQ lies above OPQ at every scale, most notably in the $G$ band (gray) and with Titan V2; the bound of Theorem~\ref{thm:bound} holds at all 160 points, with $1-\nu_k$ at 0.49 to 0.60 of it; and LAQ lowers the violation rate at $k\approx0.4n$ by 12 to 18 percent on average (10 to 20 percent per run). (g) Change in each operation across all 40 held-out runs; right: runs in which LAQ is higher. (h) LAQ (black) has the highest $G$ among seven quantizers in each of nine runs with the retrieval-tuned encoders. (i) The lowest theme agreement rises from 0.80 to 0.89.}\label{fig:samelaq}
\end{figure*}

Memories of the same size differ most at the far end. At $d/8$ bytes per item, 32 times smaller than full precision, we compare seven compressions. Bytes per item count only the per-item codes. The PCA basis, the rotation, and the codebooks are shared across all items (5.5 MB with BGE-base and 9.4 MB with BGE-large, identical for OPQ and LAQ), while rescoring additionally retains the full vectors on disk. At these bytes, a memory can store one-bit codes with or without rescoring, rotated one-bit codes, product-quantization codes, $d/8$ principal components in 8 bits, or OPQ codes with or without rescoring. Figure~\ref{fig:samelaq}a, b and Figure~\ref{fig:same} in Appendix~\ref{app:measure} compare these seven options, while Table~\ref{tab:finqa} lists all 27 compressions on FinQA with BGE-base. On FinQA with BGE-base, rescored binary codes have the highest $L$ among the first five methods (0.97) but yield a theme agreement of 0.79. PCA with 8-bit codes keeps theme agreement at 0.96. The same ordering holds with E5-base (0.96 against 0.63) and with BGE-large at 128 bytes (0.97 against 0.82). The gap also shows by company. Even when every company keeps at least 91 percent of its neighbors, rescored binary BGE-base codes move a fifth or more of the passages to another theme for 5 of the 106 FinQA companies (Figure~\ref{fig:topics}g). With PCA and 8-bit codes, every company keeps at least 86 percent of its passages in their original theme. On LoCoMo, product quantization and PCA with 8-bit codes delete a session with an agreement of 0.43 to 0.68, compared with 0.35 to 0.57 for binary codes. Across every encoder on both memories, OPQ has the highest $G$ of the seven compressions, 0.82 to 0.86 on FinQA and 0.85 to 0.91 on LoCoMo. It deletes a LoCoMo session with an agreement of 0.70 to 0.82. When rescored, OPQ reaches a top-10 agreement of 0.99 to 1.00. OPQ thus keeps both ends best among the seven compressions at these bytes, making it the natural starting point for LAQ.

%% file: t_results_laq.tex
\section{LAQ on held-out memories}\label{sec:laqresults}

LAQ raises the far end of every held-out memory, keeps the near end, and improves the far-end operations. It also has the highest $G$ among quantizers of the same size. These gains extend to an external financial memory and to a memory of one million passages.

\subsection{LAQ raises the far end of every held-out memory}

\begin{table}[t]
\centering\footnotesize
\caption{LAQ raises $G$ over its OPQ start in every held-out run. $d/8$ bytes; runs in which LAQ is higher, means over runs, and the range of the change. Retrieval-tuned encoders (BGE-base, E5-base, BGE-large), four splits each.}\label{tab:laq}
\setlength{\tabcolsep}{2.6pt}
\begin{tabular}{@{}llcccc@{}}
\toprule
Memory & & higher & OPQ & LAQ & LAQ $-$ OPQ \\
\midrule
FinQA & $G$ & 12/12 & .800 & .808 & $+.005$ to $+.010$ \\
      & $L$ & 10/12 & .861 & .864 & $-.002$ to $+.008$ \\
LoCoMo & $G$ & 12/12 & .845 & .853 & $+.005$ to $+.012$ \\
       & $L$ &  8/12 & .820 & .822 & $-.002$ to $+.009$ \\
Multi-Doc-2025 & $G$ & 12/12 & .830 & .842 & $+.007$ to $+.017$ \\
               & $L$ & 12/12 & .823 & .828 & $+.0003$ to $+.010$ \\
Multi-Doc-2025, Titan V2 & $G$ & 4/4 & .656 & .692 & $+.032$ to $+.038$ \\
                         & $L$ & 4/4 & .790 & .798 & $+.006$ to $+.011$ \\
\midrule
FinDER & $G$ & 12/12 & .821 & .835 & $+.010$ to $+.017$ \\
       & $L$ & 12/12 & .809 & .814 & $+.003$ to $+.008$ \\
\bottomrule
\end{tabular}
\end{table}

LAQ increases $G$ in all 40 held-out runs (Table~\ref{tab:laq}, Figure~\ref{fig:memgraph}h). With the retrieval-tuned encoders, the gain ranges from 0.005 to 0.017. Every gain exceeds the standard deviation of $G$ over subsamples, which is at most 0.0023. With Titan V2, whose OPQ start keeps the least of the far end, the gain reaches 0.032 to 0.038. $L$ increases in 34 of the 40 runs and stays within 0.0023 of OPQ in the rest, so the far end is gained with the near end kept.

\begin{figure*}[!p]
\centering
\includegraphics[width=\textwidth,height=0.84\textheight,keepaspectratio]{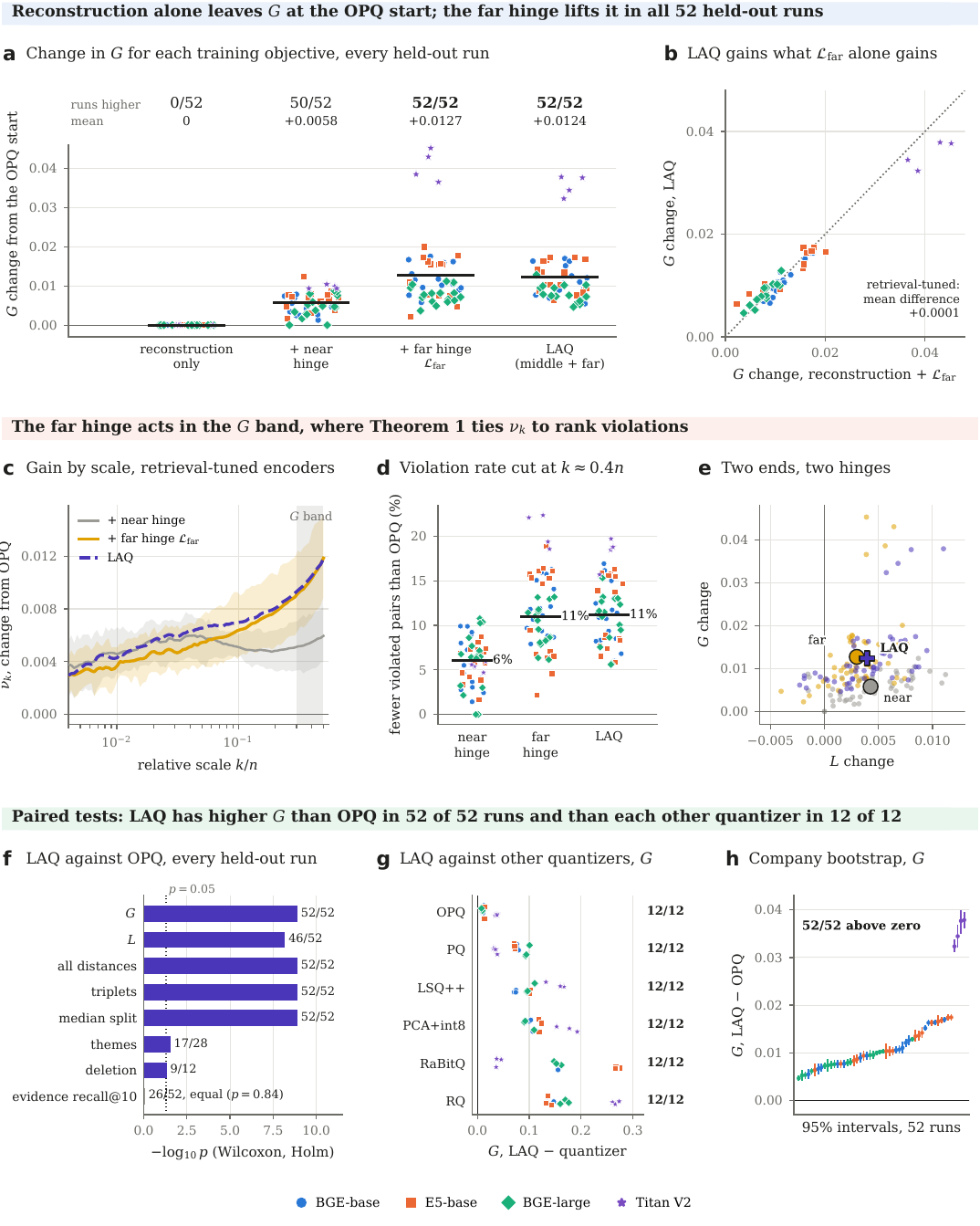}
\caption{The far hinge carries the gain in $G$, and LAQ is better than OPQ and the other quantizers in paired tests. (a) Change in $G$ from the OPQ start for each training objective on all 52 held-out runs; bars are means. (b) LAQ against reconstruction with $\mathcal L_{\mathrm{far}}$ alone, run by run. (c) Mean change in $\nu_k$ by scale with the retrieval-tuned encoders; shading, interquartile range; gray, the $G$ band. (d) Reduction of the violation rate $\bar p_k$ at $k\approx0.4n$ relative to OPQ. (e) Change in $L$ against change in $G$; large markers are means. (f) One-sided Wilcoxon tests of LAQ against OPQ over runs, Holm-corrected, with the runs in which LAQ is higher; evidence recall@10 is tested two-sided. (g) $G$ of LAQ minus that of each quantizer on Multi-Doc-2025, 12 runs. (h) Paired company-bootstrap 95 percent intervals for $G$, LAQ minus OPQ.}\label{fig:abl}
\end{figure*}

\paragraph{Per-company intervals} These gains hold across the companies of each memory. $L$, $G$, and evidence recall@10 are means over items or queries. We construct paired bootstrap intervals (95\% confidence, 10,000 resamples) by resampling these terms by held-out company. On LoCoMo we resample by session, since the held-out part of each split holds two conversations. The interval for $G$ lies above zero in all 52 held-out runs, including FinDER. The interval for $L$ lies above zero in 25 runs and below zero in none. LAQ keeps evidence recall@10: its interval contains zero in 47 of the 52 runs.

\paragraph{Mechanism} The gain spans the whole profile (Figure~\ref{fig:samelaq}c--f). Across the 16 held-out Multi-Doc-2025 memories, LAQ lifts $\nu_k$ above OPQ at all scales from $k=10$ to $k^\star$. The gain grows toward the $G$ band, where LAQ trains. Theorem~\ref{thm:bound} holds at all 160 combinations of memory, quantizer, and scale, with $1-\nu_k$ at 0.49 to 0.60 of the bound. The theorem also explains the mechanism. At $k\approx0.4n$, LAQ reduces the violation rate $\bar p_k$ by 10 to 16 percent with the retrieval-tuned encoders and by 16 to 20 percent with Titan V2. This lowers the bound on $1-\nu_k$ in the $G$ band, and the measured $\nu_k$ rises there as well.

\paragraph{Ablation} The far hinge $\mathcal L_{\mathrm{far}}$ carries the gain in $G$ (Figure~\ref{fig:abl}a, b; Table~\ref{tab:abl} in Appendix~\ref{app:runs}). We retrained every held-out run with the same recipe and checkpoint rule and changed only the terms of Eq.~\eqref{eq:J}. With reconstruction alone, training keeps the OPQ start $C_0$ in all 52 runs, since OPQ already fits its codebooks to reconstruction. A near hinge, Eq.~\eqref{eq:hingeloss} at $k=10$ or 50 on stored and ADC distances, raises $G$ by 0.0058 on average. Reconstruction with $\mathcal L_{\mathrm{far}}$ raises $G$ in all 52 runs, by 0.0127 on average, and above the near hinge in 48 of them (paired Wilcoxon test, $p=6\times10^{-9}$). The full objective raises $G$ by 0.0124. With the retrieval-tuned encoders, its $G$ equals that of $\mathcal L_{\mathrm{far}}$ alone (mean difference 0.0001), and the middle hinge adds to $L$ (0.0036 against 0.0028). With Titan V2, $\mathcal L_{\mathrm{far}}$ alone raises $G$ by 0.041 and the full objective by 0.036, with the larger gain in $L$ (0.008 against 0.005). The far end is thus learned from the order of cross pairs in the $G$ band, the pairs that Theorem~\ref{thm:bound} links to $\nu_k$.

\paragraph{Where the far hinge acts} The far hinge acts in the $G$ band (Figure~\ref{fig:abl}c--e). The near hinge lifts $\nu_k$ most near $k=0.05n$, whereas $\mathcal L_{\mathrm{far}}$ and LAQ lift it most in the $G$ band. At $k\approx0.4n$, the near hinge reduces the violation rate by 6 percent with the retrieval-tuned encoders, and $\mathcal L_{\mathrm{far}}$ and LAQ by 11 percent.

\paragraph{Paired tests} We test whether LAQ is better than OPQ with a one-sided Wilcoxon signed-rank test over the 52 held-out runs and correct each family of tests by the Holm method (Figure~\ref{fig:abl}f). Runs that share a corpus are dependent, so we repeat each test as a sign test over the 13 combinations of memory and encoder, each the mean over its splits. LAQ has higher $G$, $L$, all-distance correlation, triplet agreement, and median split at $p\le6.4\times10^{-9}$ over runs and $p=8.5\times10^{-4}$ over the 13 combinations. Theme agreement and session deletion are higher at $p=0.028$ and $p=0.046$ over runs. Evidence recall@10 is equal to that of OPQ ($p=0.84$, two-sided). The company-bootstrap intervals for $G$ lie above zero in all 52 runs (Figure~\ref{fig:abl}h).

\subsection{The gain reaches the far-end operations}

\paragraph{Global references and maps} The far-end operations of Section~\ref{sec:measure} move with $G$ (Figure~\ref{fig:samelaq}g). Both global references and the median split improve in all 40 runs. The Spearman correlation of all distances increases by 0.002 to 0.016 with the retrieval-tuned encoders and by 0.046 to 0.063 with Titan V2. The median split increases by 0.004 to 0.019 and by 0.036 to 0.043, respectively. The company map improves in 27 of 28 runs. The session map improves in 10 of 12 runs, and the remaining two stay within 0.001 of OPQ.

\paragraph{Themes and deletion} Consolidation into themes gains the most where the OPQ start keeps it least. On Multi-Doc-2025 with the retrieval-tuned encoders, the mean theme agreement rises from 0.908 to 0.931. The lowest theme agreement across the 12 runs rises from 0.800 to 0.886 (Figure~\ref{fig:samelaq}i). With Titan V2, both keep theme agreement at 0.98 or above. Across the 16 Multi-Doc-2025 runs, OPQ moves 1,203 passages to another theme, whereas LAQ moves 902. On FinQA, the mean theme agreement rises from 0.911 to 0.914. For conversational memory, deleting a LoCoMo session improves in 9 of 12 runs, with mean agreement rising from 0.686 to 0.695.

\paragraph{Retrieval} Retrieval is served by the near end, which LAQ keeps. Top-10 agreement increases in 27 of 40 runs. Mean evidence recall@10 stays within 0.005 of OPQ in every memory: 0.545 versus 0.546 on FinQA, 0.649 versus 0.649 on LoCoMo, 0.795 versus 0.796 on Multi-Doc-2025, and 0.749 versus 0.754 with Titan V2. Peer groups and the information barrier are nearest-neighbor questions among company centroids. As Section~\ref{sec:measure} found, they follow $L$ at least as closely as $G$. LAQ raises peer agreement in 17 of 28 runs and keeps barrier agreement within 0.011 of OPQ in every run. All runs are listed in Appendix~\ref{app:runs} (Table~\ref{tab:runs}).

\subsection{LAQ against other quantizers at the same size}

\begin{table}[t]
\centering\footnotesize
\caption{LAQ has the highest $G$ with every encoder. $G$ / $L$ on held-out Multi-Doc-2025 at $d/8$ bytes, means over the first three splits. RaBitQ uses 104 and 136 bytes. ScaNN is fitted on the held-out memory and searched with full-precision queries; we report its $L$.}\label{tab:quant}
\setlength{\tabcolsep}{2.2pt}
\begin{tabular}{@{}lcccc@{}}
\toprule
 & BGE-base & E5-base & BGE-large & Titan V2 \\
\midrule
LAQ   & \textbf{.832} / \textbf{.817} & \textbf{.830} / \textbf{.807} & \textbf{.866} / \textbf{.857} & \textbf{.696} / .800 \\
OPQ   & .820 / .814 & .816 / .801 & .857 / .852 & .659 / .792 \\
PQ    & .754 / .743 & .759 / .749 & .769 / .771 & .661 / .787 \\
LSQ++ & .759 / .725 & .730 / .689 & .764 / .744 & .543 / .679 \\
RQ    & .693 / .582 & .693 / .552 & .697 / .598 & .429 / .511 \\
PCA + int8 & .729 / .631 & .710 / .584 & .768 / .695 & .521 / .578 \\
RaBitQ & .677 / .666 & .561 / .545 & .712 / .704 & .656 / .776 \\
ScaNN & -- / .615   & -- / .507   & -- / .578   & -- / \textbf{.837} \\
\bottomrule
\end{tabular}
\end{table}

LAQ has the highest $G$ among the quantizers of its size (Table~\ref{tab:quant}, Figure~\ref{fig:samelaq}h). In each of the nine runs with the retrieval-tuned encoders, LAQ has both the highest $G$ and the highest $L$ of all codes without rescoring. LSQ++, the strongest additive quantizer tested, reaches a $G$ of 0.73 to 0.77, against 0.83 to 0.87 for LAQ. With Titan V2, LAQ has the highest $G$ in all three runs, and ScaNN, fitted on the held-out memory itself, has the highest $L$. With rescoring of the 40 nearest codes, LAQ reaches an $L$ of 0.995 to 0.999 in all twelve runs. This is above rescored sign codes (0.861 to 0.996) and rescored ITQ codes (0.938 to 0.994) in each run, and the stored LAQ codes keep the highest $G$. Against each of OPQ, PQ, LSQ++, RQ, PCA + int8, and RaBitQ, LAQ has higher $G$ and $L$ in all 12 runs (paired Wilcoxon test, $p=2.9\times10^{-3}$ after Holm correction; Figure~\ref{fig:abl}g).

\subsection{An external financial memory}

These gains hold on FinDER, a memory built by others from the queries of financial professionals. It was run once, with the recipe unchanged, on 124 held-out companies in each of four splits (Table~\ref{tab:laq}). LAQ raises $G$ in all 12 runs by 0.010 to 0.017, lifting the mean from 0.821 to 0.835. Every gain exceeds the standard deviation of $G$ over subsamples, which is at most 0.0025 here. $L$ increases in all 12 runs by 0.003 to 0.008, and top-10 agreement rises in all 12. Mean evidence recall@10 is 0.458 for both, within 0.007 of OPQ in every run.

\subsection{A memory of a million passages}\label{sec:marco}

LAQ also raises $G$ on a memory of one million passages. The memory holds 1,000,000 passages of MS MARCO~\citep{bajaj2016msmarco}, encoded with BGE-base and stored at 96 bytes each. OPQ and LAQ were fitted on 100,000 other passages, and the 6,980 development queries search the memory by ADC. At this size, training draws a new subsample of 8,192 reference items at each step and chooses the cross pairs by their ranks within it, the argument behind the estimator of $G$. It decodes only the items a step uses and averages the reconstruction term over 16,384 items. All other settings are those of the recipe. Over three seeds, LAQ raises $G$ from 0.774 to 0.784 to 0.785. The gain of 0.010 to 0.011 is about three times the standard deviation of $G$ over subsamples (at most 0.0039). $L$, computed on the 10 nearest neighbors of 20,000 random passages, stays within 0.002 of OPQ. Top-10 agreement is 0.818 for OPQ and 0.818 to 0.820 for LAQ. Evidence recall@10, over the 820 queries whose relevant passage is in the memory, is 0.860 for OPQ and within 0.008 of it for LAQ. One training run takes 175 seconds and at most 10.0 GiB of one 48~GB GPU, and a second run reproduces the codebooks bit for bit.

%% file: t_discussion.tex
\section{Discussion}\label{sec:discussion}

\paragraph{Why the far end can be learned} Ordering cross pairs controls the far end, and this is what makes LAQ work. The bound has a leading factor near 2 across the $G$ band and near $\sqrt{N/10}$ at $k=10$. Reconstruction and nearest-neighbor objectives serve the near end. LAQ combines the two through a hinge loss at middle and far scales, a normalized reconstruction term for the near end, and a checkpoint selected on the far end of a monitor set. The ablation of Section~\ref{sec:laqresults} isolates the far hinge: reconstruction alone keeps the OPQ start, and $\mathcal L_{\mathrm{far}}$ alone gives the full gain in $G$ in all 52 runs (Figure~\ref{fig:abl}). The gain is largest where the OPQ codes keep the least of the far end, as with Titan V2, and it reaches the operations that follow $G$.

\paragraph{Cost against other quantizers} LAQ trains in minutes, between PQ and the additive quantizers, and keeps the query cost of OPQ. Each LAQ step costs about one $k$-means iteration of PQ and OPQ~\citep{jegou2011pq,ge2014opq}, plus a one-time ranking of the training memory. In contrast, additive quantizers spend their cost on encoding. RQ keeps several partial codes per item in a beam search, and LSQ++ encodes vectors by joint local search over all codebooks~\citep{chen2010rvq,martinez2018lsq}. ITQ alternates a $d\times d$ rotation with sign codes~\citep{gong2013itq}, and RaBitQ applies a random rotation without training~\citep{gao2024rabitq}. On the Multi-Doc-2025 training memories, the faiss implementations~\citep{douze2024faiss} on CPU took 6 to 12 seconds to fit PQ, 146 to 517 seconds for ITQ, 511 to 1,193 seconds for RQ, and 8.2 to 17.1 hours for LSQ++. By comparison, LAQ took 35 to 112 seconds on one GPU after its OPQ start. LAQ and OPQ encode a new item in $O(Kd)$ time, while RQ and LSQ++ repeat their search for every item added to the memory. LAQ thus has the highest $G$ among these quantizers (Table~\ref{tab:quant}) at a training cost between PQ and the additive quantizers.

\paragraph{What to report} A compressed memory can be reported with $(L,G)$ against the memory it replaces, computed on the operator's own vectors. $L$ says whether queries return the same items. $G$ says whether the memory keeps its far end: which items are consolidated or deleted together, and how sessions, companies, and themes relate. Neither needs labels, and $G$ is estimated from a subsample of 2,000 items. Where a memory serves consolidation or deletion, $G$ belongs in its acceptance test. The compression is then chosen as the one with the highest $G$ among those with the required $L$ at the bytes available. Theorem~\ref{thm:bound} adds a diagnostic: the violation rate of cross pairs at a scale shows which part of the profile a compression loses.

\paragraph{Financial memory} The consequences are clearest in financial memories. A consolidation that groups filings differently changes the sector briefs an analyst agent writes. A deletion that selects different passages changes what a client request removes. Peer groups, the information barrier, and the company map compare company centroids. A check of these operations combines $L$ with agreement measured on the firm's own companies.

\paragraph{Extensions} Residual and additive codebooks are natural next targets for this objective, and sampling reference items already extends training to one million passages (Section~\ref{sec:marco}).

\section{Conclusion}

Agent memories are compressed and accepted by their nearest neighbors, even though agents also consolidate, delete, and map what they hold. One label-free profile measures both ends. Its near end predicts retrieval, and with retrieval-tuned encoders its far end predicts consolidation and deletion, each beyond the other. Our main result is that the order of near and far items bounds the profile at every scale, and that this bound can be trained. Locality-aware quantization (LAQ) learns this order and raises the global fidelity of every held-out memory at the same bytes and search as the OPQ index it replaces. To our knowledge, LAQ is the first quantizer whose objective comes from a proven bound on the far end of a memory. It needs no labels and no change at query time, and it applies wherever embeddings are compressed and then clustered, deduplicated, or mapped. For agent memory, $(L,G)$ gives an acceptance test for both ends, and LAQ gives a way to pass it.

Data and code: {\color{blue}\url{https://github.com/hank08819/LAQ}}

%% file: t_appendix_proofs.tex
\section{Proofs}\label{app:proofs}

This appendix proves the results of Sections~\ref{sec:profile} and~\ref{sec:laq}, except Theorem~\ref{thm:bound}, whose proof is in Section~\ref{sec:bound}.

\begin{proof}[Proof of Proposition~\ref{prop:chance}]
(i) Each of the $k$ members of $\mathcal N^x_k(i)$ lies in a uniform $k$-subset of $N$ items with probability $k/N$, so $\mathbb E|\mathcal N^x_k(i)\cap\mathcal N^z_k(i)|=k^2/N$. Substitution in~\eqref{eq:nu} gives $0$. (ii) A rotation, reflection, or translation keeps every distance, and a uniform scaling multiplies every distance by one positive constant. A strictly increasing transformation of a distance keeps its order. In each case every ranking $r^z_i$, and hence every $\mathcal N^z_k(i)$, is unchanged.
\end{proof}

\begin{proof}[Proof of Proposition~\ref{prop:duality}]
Inclusion--exclusion gives the identity. At $2k^\star=N$ it reads $|\mathcal F^x\cap\mathcal F^z|=|\mathcal N^x\cap\mathcal N^z|$, so item $i$ has $2|\mathcal N^x_{k^\star}(i)\cap\mathcal N^z_{k^\star}(i)|$ agreements out of $N$ and $A=\DLP_{k^\star}$. Equation~\eqref{eq:nu} with $N-k^\star=k^\star$ gives $\nu_{k^\star}=2\DLP_{k^\star}-1$.
\end{proof}

\begin{proof}[Proof of Proposition~\ref{prop:sheppard}]
The transformations map both medians to $0$. For a bivariate normal pair with unit variances and correlation $\gamma$, Sheppard's formula~\citep{sheppard1899} gives $\Pr(\text{both coordinates on the same side of }0)=\frac12+\frac1\pi\arcsin \gamma$, so $\mathbb E(2A_i-1)=\frac2\pi\arcsin \gamma_i$. Kendall's $\tau$ and Spearman's $\rho$ of the bivariate normal take the forms above~\citep{kruskal1958}, and both increase in $\gamma$.
\end{proof}

\begin{proof}[Proof of Corollary~\ref{cor:hinge}]
A violated pair has $\Delta^z\le0$, so $h\ge\eta\Delta^x$. If also $\Delta^x\ge\delta$, then $h\ge\eta\delta$. Hence $\mathbb 1[\text{violated}]\le\mathbb 1[\Delta^x<\delta]+h/(\eta\delta)$, and taking expectations gives~\eqref{eq:hinge}. Rankings by squared distance equal rankings by distance, so $\Delta^z\le0$ for every violated pair, with equality when two compressed distances tie.
\end{proof}

\begin{proof}[Proof of Proposition~\ref{prop:noise}]
As $n\to\infty$, the $k$ nearest items of $i$ are those with $u_j<z_\alpha$, so a cross pair $(p,q)$ has $v=u_p<z_\alpha<u=u_q$, with density $\phi(u)\phi(v)/(\alpha(1-\alpha))$ on this region. The pair is violated when $\tilde D_q<\tilde D_p$, that is, when $u+e_q/s_i<v+e_p/s_i$. The difference $(e_p-e_q)/s_i$ is normal with variance $2\omega_i^2$, so this happens with probability $\Phi(-(u-v)/(\sqrt2\,\omega_i))$. Integrating over the region gives~\eqref{eq:noise}. For small $\omega$, write $u=z_\alpha+x$ and $v=z_\alpha-y$ with $x,y\ge0$. The factor $\Phi(-(x+y)/(\sqrt2\,\omega))$ confines the integral to $x+y=O(\omega)$, where $\phi(u)\phi(v)=\phi(z_\alpha)^2(1+O(x+y))$. With $t=x+y$, the area element of $\{x,y\ge0,\ x+y\in dt\}$ is $t\,dt$, and
\[
\int_0^\infty t\,\Phi\Bigl(-\frac{t}{\sqrt2\,\omega}\Bigr)dt=2\omega^2\int_0^\infty s\,\Phi(-s)\,ds=\frac{\omega^2}{2},
\]
since $\int_0^\infty s\,\Phi(-s)\,ds=\frac12\int_0^\infty s^2\phi(s)\,ds=\frac14$ by parts. The $O(x+y)$ term contributes $O(\omega^3)$.
\end{proof}

\begin{proof}[Proof of Corollary~\ref{cor:noiseprofile}]
Write $\epsilon_j=e_j/s_i$. The compressed distances rank the items as $u_j+\epsilon_j$ does, and $y_j=(u_j+\epsilon_j)/\sqrt{1+\omega_i^2}$ is $\mathcal N(0,1)$ with correlation $\varrho_i$ to $u_j$. As $n\to\infty$, the $k$ nearest items of $i$ are those with $u_j<z_\alpha$ before compression and those with $y_j<z_\alpha$ after it, so the share of the $k$ nearest items that both keep converges to $\Phi_2(z_\alpha,z_\alpha;\varrho_i)/\alpha$. Substituting this share for $\DLP_k$ in~\eqref{eq:nu}, with $k/N\to\alpha$, gives~\eqref{eq:noiseprofile}. Plackett's identity $\partial\Phi_2(z,z;\varrho)/\partial\varrho=\phi_2(z,z;\varrho)>0$, with $\phi_2$ the bivariate normal density, shows that $\nu$ increases strictly in $\varrho_i$, which decreases strictly in $\omega_i$. At $\alpha=1/2$, $z_\alpha=0$ and Sheppard's formula gives $\Phi_2(0,0;\varrho)=\frac14+\frac{1}{2\pi}\arcsin\varrho$, so $\nu=\frac2\pi\arcsin\varrho_i$. Finally, $1-\nu=(\alpha-\Phi_2)/(\alpha(1-\alpha))$, and $\alpha-\Phi_2$ is the probability that $u_j<z_\alpha<u_j+\epsilon_j$ after rescaling. As $\omega_i\to0$, this probability is $\phi(z_\alpha)\,\mathbb E[\epsilon_j]_+ +o(\omega_i)=\phi(z_\alpha)\,\omega_i/\sqrt{2\pi}+o(\omega_i)$, so $1-\nu=\phi(z_\alpha)\,\omega_i/\bigl(\sqrt{2\pi}\,\alpha(1-\alpha)\bigr)+o(\omega_i)$, which is $0.64\,\omega_i$ at $\alpha=0.4$.
\end{proof}

\begin{proof}[Proof of Proposition~\ref{prop:start}]
Because no subvector is equidistant from its two nearest codewords, the assignments of $C_0$ stay unchanged for small $\epsilon$, and every decoded vector is then a linear function of the codebooks. The reconstruction error is $\mathcal L_{\mathrm{rec}}(C)=\frac1n\sum_{\ell,j}\sum_{i:\,a_\ell(i)=j}\|y_{i,\ell}-c_{\ell,j}\|^2$, with $a_\ell(i)$ the assignment, a quadratic in $C$. Its gradient with respect to $c_{\ell,j}$ is $\frac{2}{n}\sum_{i:\,a_\ell(i)=j}(c_{\ell,j}-y_{i,\ell})$, which is zero when $c_{\ell,j}$ is the mean of its subvectors, and its Hessian is $\frac{2n_{\ell j}}{n}$ times the identity on $c_{\ell,j}$. A second-order expansion along $-h$, which has no remainder for a quadratic, gives the first line of~\eqref{eq:start}. The second line is the first-order expansion of $\mathcal L_{\mathrm{far}}$ along $-h$, whose directional derivative is $-\langle g,h\rangle$.
\end{proof}

%% file: t_appendix.tex
\section{LAQ recipe}\label{app:recipe}

Every LAQ run uses one recipe: $M=d/8$ subspaces of 8 dimensions with 256 codewords each, the full PCA basis, 1,000 steps, 1,024 anchors per scale group and step, $\lambda_{\mathrm{mid}}=\lambda_{\mathrm{far}}=1$, margin $\eta=1$, Adam with a learning rate of $0.002$ times the standard deviation of $C_0$ and cosine decay, temperature from $T_0$ to $0.05\,T_0$, a monitor set of 2,000 training items checked every 250 steps, and a reconstruction limit of 1.15 for a checkpoint. The scale normalizers $s_{\mathrm{mid}}$ and $s_{\mathrm{far}}$ are medians of 16,384 uncompressed gaps sampled at the start.

%% file: t_appendix_d.tex
\section{Additional measurement results}\label{app:measure}

\begin{figure}[H]
\centering
\includegraphics[width=\textwidth]{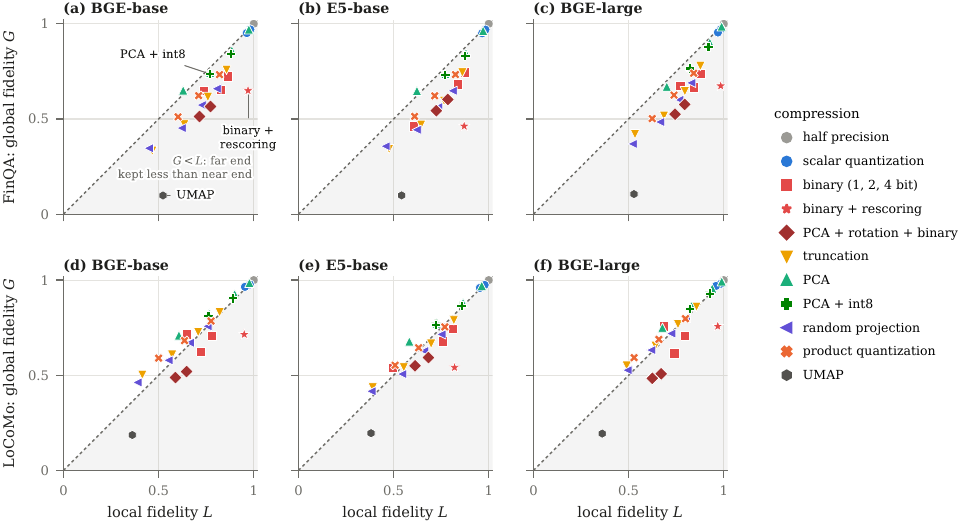}
\caption{Compressions of one memory spread off a single curve, and rescored binary codes lie furthest below the diagonal, by 0.21 to 0.41. $L$ and $G$ of 27 compressions of the two agent memories, three retrieval-tuned encoders; Figure~\ref{fig:otherenc} gives the other three. Dashed line: $G=L$; shaded: less of the far end kept than of the near end.}\label{fig:plane}
\end{figure}

\begin{figure}[H]
\centering
\includegraphics[width=\textwidth]{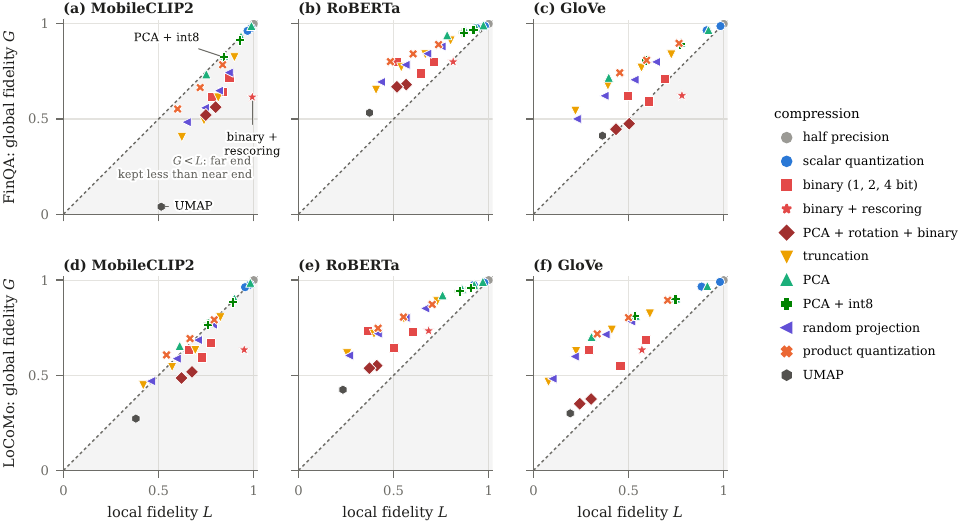}
\caption{With encoders of other kinds, PCA and product quantization lie on or above the diagonal, and one-bit codes of GloVe and RoBERTa keep more of the far end than of the near end. $L$ and $G$ of 27 compressions of the two agent memories; dashed line: $G=L$.}\label{fig:otherenc}
\end{figure}

\begin{figure}[H]
\centering
\includegraphics[width=\textwidth]{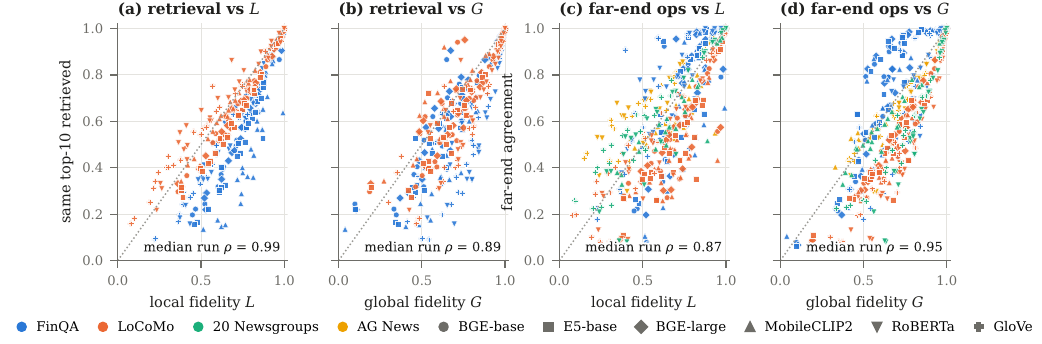}
\caption{Retrieval follows $L$ ($\rho$ 0.99, against 0.89 with $G$), and the far-end operations follow $G$ ($\rho$ 0.95, against 0.87 with $L$). Agreement with the uncompressed memory, every compression and run. (a, b) Retrieval on the two agent memories. (c, d) Consolidation into themes for FinQA, deletion for LoCoMo and the topic corpora. $\rho$: median over runs of the Spearman correlation across the 27 compressions of a run.}\label{fig:predict}
\end{figure}

\begin{figure}[H]
\centering
\includegraphics[width=\textwidth]{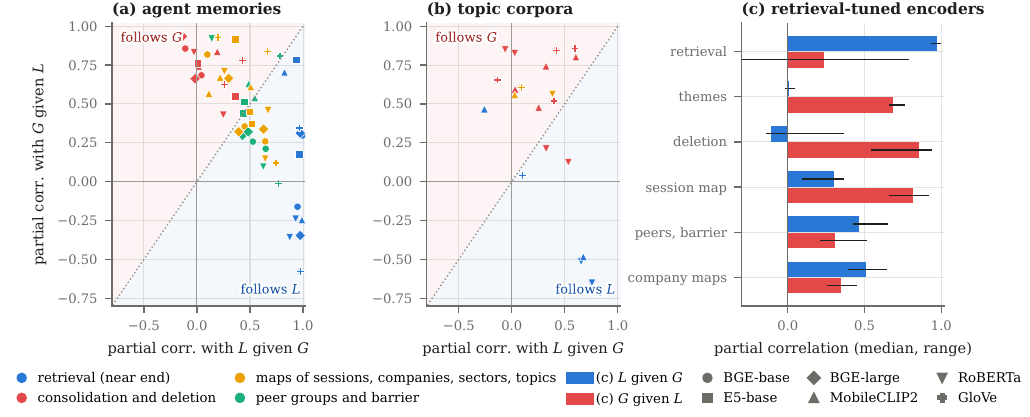}
\caption{Each operation follows the end at which it compares items: retrieval lies in the $L$ region, consolidation and deletion in the $G$ region. Partial rank correlation with $G$ given $L$ against that with $L$ given $G$, across the 27 compressions of each run. (a) Agent memories, (b) topic corpora; color: operation; shape: encoder. (c) Medians and ranges over the six runs of the retrieval-tuned encoders.}\label{fig:partial}
\end{figure}

\begin{figure}[H]
\centering
\includegraphics[width=\textwidth]{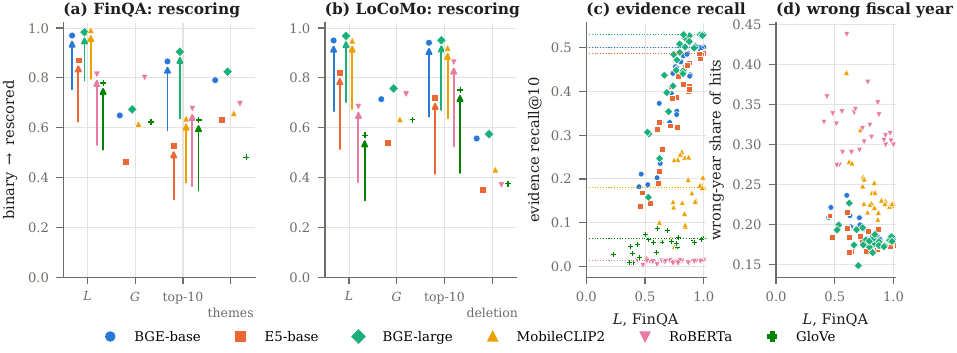}
\caption{Rescoring raises $L$ and retrieval and leaves $G$ and the far-end operations at the values of the binary codes. (a, b) Binary codes before and after full-precision rescoring of the 40 nearest codes; arrows run from binary to rescored, and a marker without an arrow is unchanged. (c) Evidence recall@10 on FinQA rises with $L$; dotted lines: uncompressed memories. (d) The share of right-company passages from another fiscal year stays near its uncompressed level.}\label{fig:rescore}
\end{figure}

\begin{figure}[H]
\centering
\includegraphics[width=\textwidth]{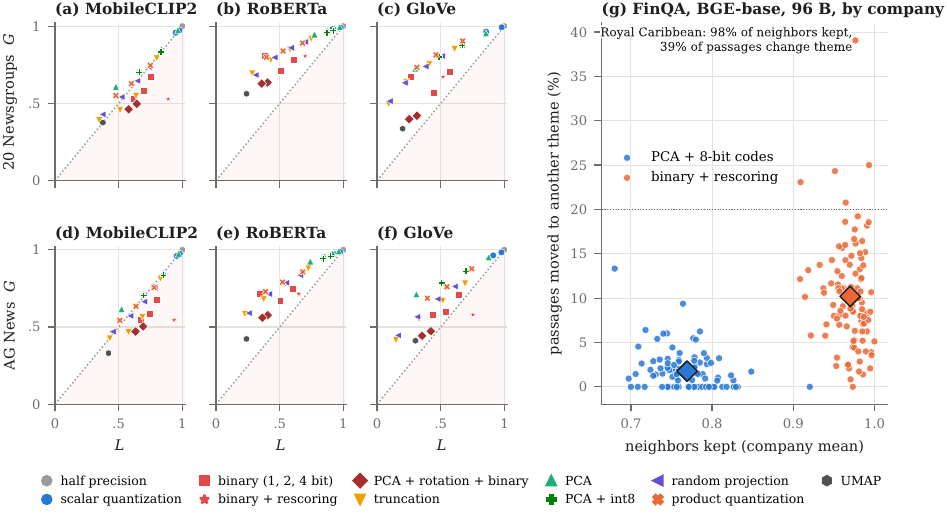}
\caption{On the topic corpora, PCA and product quantization lie on or above the diagonal (a--f). On FinQA, rescored binary codes keep neighbors yet move up to 39 percent of a company's passages to another theme (g). (a--f) $L$ and $G$ of 27 compressions, three encoders of other kinds; dotted line: $G=L$. (g) The two 96-byte compressions of FinQA (BGE-base) by company; diamonds: means; dotted line: 20 percent moved.}\label{fig:topics}
\end{figure}

\begin{figure}[H]
\centering
\includegraphics[width=\textwidth]{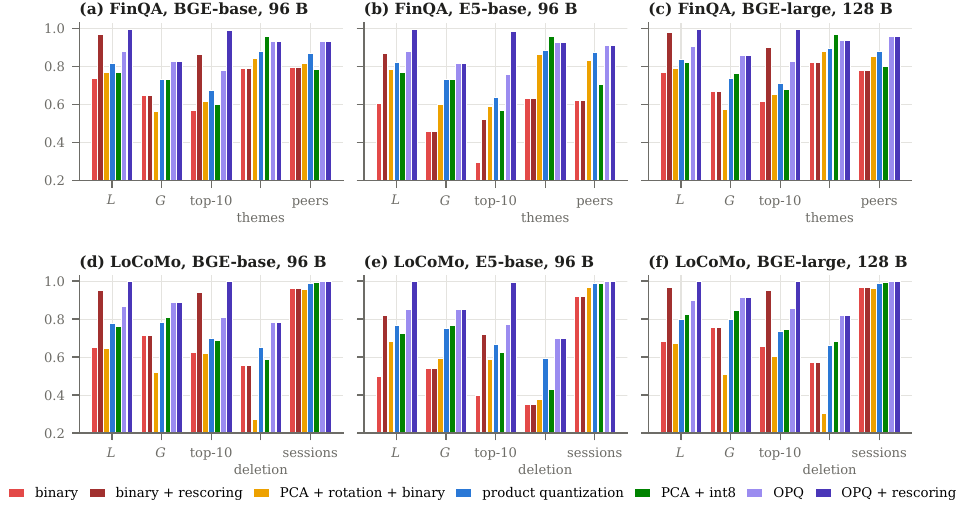}
\caption{At $d/8$ bytes, OPQ keeps the highest $G$ of seven compressions, and rescored binary codes keep a high $L$ with a lower $G$. Fidelity and agreement with the uncompressed memory, retrieval-tuned encoders; (a--c) FinQA, (d--f) LoCoMo.}\label{fig:same}
\end{figure}

The figures and tables of this appendix give every run behind the main figures. Figure~\ref{fig:plane} places the 27 compressions of the two agent memories on the $(L,G)$ plane for the retrieval-tuned encoders, and Figure~\ref{fig:otherenc} for the three encoders of other kinds. Figures~\ref{fig:predict} and~\ref{fig:partial} relate each operation to $L$ and $G$, and Figure~\ref{fig:rescore} gives rescoring. The topic corpora repeat the pattern of the agent memories: Figure~\ref{fig:topics} places their compressions on the plane and breaks the 96-byte comparison of FinQA down by company. Figure~\ref{fig:same} gives the seven compressions at $d/8$ bytes. Tables~\ref{tab:nonlinear}--\ref{tab:finqa} give the nonlinear maps, the correlations behind Figures~\ref{fig:predict} and~\ref{fig:partial}, rescoring for every code family that keeps Euclidean distances, and every compression of the financial memory.

\FloatBarrier
\begin{figure}[H]
\centering
\includegraphics[width=\textwidth]{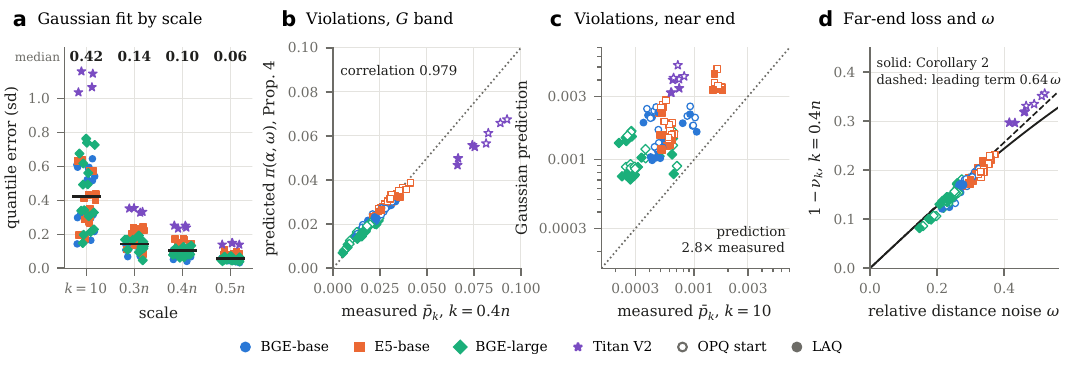}
\caption{The normal model of Proposition~\ref{prop:noise} holds in the $G$ band and predicts its violations and its loss of profile. All 52 held-out runs; open markers, OPQ start; filled, LAQ. (a) Median error of each anchor's standardized distance quantile against the normal quantile, in standard deviations, at the near end and in the $G$ band. (b) The full integral~\eqref{eq:noise} against the measured violation rate at $k=0.4n$. (c) The leading term at $k=10$ against the measured rate. (d) Measured $1-\nu_k$ at $k=0.4n$ against the mean relative distance noise $\omega$; solid line, Corollary~\ref{cor:noiseprofile}; dashed line, its leading term $0.64\,\omega$ (Appendix~\ref{app:proofs}).}\label{fig:noise}
\end{figure}

\FloatBarrier
\begin{table}[H]
\centering\normalsize\renewcommand{\arraystretch}{1.12}
\setlength{\tabcolsep}{6pt}
\caption{Neighborhood-embedding maps keep $G$ at 0.22 or less, and nonnegative PCA keeps more of the far end than of the near end. Maps fitted to the memory (BGE-base), kept in full precision; PCA to 16 dimensions is the linear reference for NPCA. Queries are mapped by the fitted transform (UMAP~\citep{mcinnes2018}, PHATE~\citep{moon2019}, Isomap~\citep{tenenbaum2000}, NPCA~\citep{han2010tcbb}, PCA) or placed at the inverse-distance-weighted mean of the positions of their 10 nearest stored items (t-SNE~\citep{vandermaaten2008}, SR-tSNE~\citep{han2026srtsne}). Top-10 agreement: overlap of the top-10 lists. Last column: theme agreement (FinQA) or agreement of session deletion (LoCoMo).}\label{tab:nonlinear}
\begin{tabular}{@{}llrcccc@{}}
\toprule
Memory & Map & Bytes & $L$ & $G$ & Retrieval & Themes / deletion \\
\midrule
\input{gen/tab_nonlinear_rows} 
\bottomrule
\end{tabular}
\end{table}

\begin{table}[H]
\centering\normalsize\renewcommand{\arraystretch}{1.12}
\setlength{\tabcolsep}{6pt}
\caption{Retrieval follows $L$, and consolidation and deletion follow $G$. Spearman correlation across 27 compressions of each operation's agreement with the uncompressed memory (label precision for the topic corpora) with $L$ and with $G$, and the partial rank correlation with each end given the other ($L\mid G$: with $L$ given $G$; $G\mid L$: with $G$ given $L$). Agent memories, FinQA.}\label{tab:pred}
\begin{tabular}{@{}llcccc@{\hspace{10pt}}c@{}}
\toprule
Memory, encoder & Operation & End & $L$ & $G$ & $L\mid G$ & $G\mid L$ \\
\midrule
\input{gen/tab_pred_a} 
\bottomrule
\end{tabular}
\end{table}

\begin{table}[H]
\centering\normalsize\renewcommand{\arraystretch}{1.12}
\setlength{\tabcolsep}{6pt}
\caption{The same split holds on LoCoMo and the topic corpora. Continuation of Table~\ref{tab:pred}.}\label{tab:predb}
\begin{tabular}{@{}llcccc@{\hspace{10pt}}c@{}}
\toprule
Memory, encoder & Operation & End & $L$ & $G$ & $L\mid G$ & $G\mid L$ \\
\midrule
\input{gen/tab_pred_b} 
\bottomrule
\end{tabular}
\end{table}

\begin{table}[H]
\centering\normalsize\renewcommand{\arraystretch}{1.12}
\setlength{\tabcolsep}{6pt}
\caption{Rescoring raises $L$ and retrieval for every code family, and $G$ and the far-end operations keep the values of the stored codes. Code families that keep Euclidean distances, $d/8$ bytes per item. The 40 nearest codes are rescored with the full-precision vectors; $L$ and retrieval come from the rescored lists, $G$ and consolidation or deletion from the stored codes. Top-10 agreement: overlap of the top-10 lists. Last column: theme agreement (FinQA) or agreement of session deletion (LoCoMo).}\label{tab:rescoreall}
\begin{tabular}{@{}lllcccc@{}}
\toprule
Memory & Encoder & Compression & $L$ & $G$ & Retrieval & Themes / deletion \\
\midrule
\input{gen/tab_rescore_rows} 
\bottomrule
\end{tabular}
\end{table}

\begin{table}[H]
\centering\small\renewcommand{\arraystretch}{1.3}
\setlength{\tabcolsep}{5pt}
\caption{At equal bytes, compressions of the financial memory differ most in $G$ and theme agreement. FinQA, BGE-base, 27 compressions: bytes per item, fidelity, and agreement with the uncompressed memory. Top-10 agreement: overlap of the top-10 lists for 863 questions. Evidence: evidence recall@10 (0.50 uncompressed). Themes: theme agreement over 11 themes. Peers: overlap of five-company peer groups. Barrier: Jaccard overlap of the memories removed to wall off a company. Map: rank correlation of distances among company centroids.}\label{tab:finqa}
\begin{tabular}{@{}lrcccccccc@{}}
\toprule
Compression & Bytes & $L$ & $G$ & Retrieval & Evidence & Themes & Peers & Barrier & Map \\
\midrule
\input{gen/tab_finqa_rows} \bottomrule
\end{tabular}
\end{table}
\FloatBarrier

%% file: gen/tab_nonlinear_rows.tex
FinQA & UMAP, 16 dims & 64 & 0.523 & 0.10 & 0.24 & 0.08 \\
 & PCA to 50, UMAP to 16 & 64 & 0.474 & 0.18 & 0.16 & 0.21 \\
 & PHATE, 16 dims & 64 & 0.537 & 0.15 & 0.19 & 0.08 \\
 & UMAP, 2 dims & 8 & 0.453 & 0.07 & 0.18 & 0.07 \\
 & PHATE, 2 dims & 8 & 0.231 & 0.12 & 0.04 & 0.07 \\
 & t-SNE, 2 dims & 8 & 0.585 & 0.13 & 0.07 & 0.09 \\
 & SR-tSNE, 2 dims & 8 & 0.590 & 0.13 & 0.07 & 0.09 \\
 & Isomap, 16 dims & 64 & 0.560 & 0.25 & 0.28 & 0.15 \\
 & NPCA, 16 dims & 64 & 0.237 & 0.43 & 0.10 & 0.41 \\
 & PCA, 16 dims (linear) & 64 & 0.349 & 0.54 & 0.17 & 0.71 \\
\addlinespace[2pt]
LoCoMo & UMAP, 16 dims & 64 & 0.362 & 0.19 & 0.30 & 0.08 \\
 & PCA to 50, UMAP to 16 & 64 & 0.329 & 0.19 & 0.25 & 0.07 \\
 & PHATE, 16 dims & 64 & 0.292 & 0.22 & 0.24 & 0.08 \\
 & UMAP, 2 dims & 8 & 0.255 & 0.16 & 0.21 & 0.04 \\
 & PHATE, 2 dims & 8 & 0.141 & 0.19 & 0.14 & 0.05 \\
 & t-SNE, 2 dims & 8 & 0.388 & 0.20 & 0.15 & 0.05 \\
 & SR-tSNE, 2 dims & 8 & 0.392 & 0.20 & 0.15 & 0.05 \\
 & Isomap, 16 dims & 64 & 0.388 & 0.43 & 0.33 & 0.23 \\
 & NPCA, 16 dims & 64 & 0.220 & 0.44 & 0.18 & 0.11 \\
 & PCA, 16 dims (linear) & 64 & 0.349 & 0.54 & 0.26 & 0.14 \\

%% file: gen/tab_pred_a.tex
FinQA, BGE-base & same top-10 retrieved & near & 1.00 & 0.89 & 0.99 & 0.30 \\
 & same themes (ARI) & far & 0.80 & 0.90 & 0.04 & 0.68 \\
 & same peer groups & far & 0.87 & 0.83 & 0.53 & 0.26 \\
 & information barrier & far & 0.91 & 0.84 & 0.65 & 0.21 \\
 & company map & far & 0.91 & 0.86 & 0.64 & 0.26 \\
 & sector map & far & 0.86 & 0.84 & 0.45 & 0.35 \\
\addlinespace[2pt]
FinQA, E5-base & same top-10 retrieved & near & 0.98 & 0.94 & 0.94 & 0.78 \\
 & same themes (ARI) & far & 0.83 & 0.93 & 0.01 & 0.76 \\
 & same peer groups & far & 0.89 & 0.90 & 0.45 & 0.51 \\
 & information barrier & far & 0.88 & 0.88 & 0.44 & 0.44 \\
 & company map & far & 0.90 & 0.89 & 0.50 & 0.45 \\
 & sector map & far & 0.89 & 0.87 & 0.52 & 0.37 \\
\addlinespace[2pt]
FinQA, BGE-large & same top-10 retrieved & near & 1.00 & 0.92 & 0.97 & 0.31 \\
 & same themes (ARI) & far & 0.83 & 0.91 & -0.02 & 0.66 \\
 & same peer groups & far & 0.88 & 0.86 & 0.43 & 0.30 \\
 & information barrier & far & 0.90 & 0.88 & 0.48 & 0.32 \\
 & company map & far & 0.94 & 0.91 & 0.63 & 0.34 \\
 & sector map & far & 0.87 & 0.86 & 0.39 & 0.32 \\
\addlinespace[2pt]
FinQA, MobileCLIP2 & same top-10 retrieved & near & 0.95 & 0.92 & 0.83 & 0.70 \\
 & same themes (ARI) & far & 0.77 & 0.90 & 0.02 & 0.74 \\
 & same peer groups & far & 0.88 & 0.90 & 0.49 & 0.63 \\
 & information barrier & far & 0.88 & 0.88 & 0.55 & 0.54 \\
 & company map & far & 0.88 & 0.90 & 0.51 & 0.61 \\
 & sector map & far & 0.73 & 0.83 & 0.11 & 0.57 \\
\addlinespace[2pt]
FinQA, RoBERTa & same top-10 retrieved & near & 0.98 & 0.88 & 0.93 & -0.24 \\
 & same themes (ARI) & far & 0.84 & 0.86 & 0.25 & 0.43 \\
 & same peer groups & far & 0.91 & 0.85 & 0.63 & 0.10 \\
 & information barrier & far & 0.91 & 0.99 & 0.14 & 0.92 \\
 & company map & far & 0.95 & 0.93 & 0.67 & 0.46 \\
 & sector map & far & 0.92 & 0.87 & 0.64 & 0.15 \\
\addlinespace[2pt]
FinQA, GloVe & same top-10 retrieved & near & 0.99 & 0.83 & 0.97 & 0.35 \\
 & same themes (ARI) & far & 0.77 & 0.86 & 0.26 & 0.63 \\
 & same peer groups & far & 0.90 & 0.72 & 0.77 & -0.01 \\
 & information barrier & far & 0.92 & 0.93 & 0.78 & 0.81 \\
 & company map & far & 0.89 & 0.94 & 0.66 & 0.84 \\
 & sector map & far & 0.90 & 0.76 & 0.74 & 0.12 \\
\addlinespace[2pt]

%% file: gen/tab_pred_b.tex
LoCoMo, BGE-base & same top-10 retrieved & near & 0.99 & 0.89 & 0.95 & -0.16 \\
 & same memories deleted & far & 0.87 & 0.97 & -0.11 & 0.86 \\
 & session map & far & 0.89 & 0.97 & 0.10 & 0.82 \\
\addlinespace[2pt]
LoCoMo, E5-base & same top-10 retrieved & near & 0.99 & 0.89 & 0.96 & 0.18 \\
 & same memories deleted & far & 0.87 & 0.90 & 0.36 & 0.55 \\
 & session map & far & 0.90 & 0.98 & 0.36 & 0.92 \\
\addlinespace[2pt]
LoCoMo, BGE-large & same top-10 retrieved & near & 0.99 & 0.90 & 0.97 & -0.35 \\
 & same memories deleted & far & 0.90 & 0.99 & -0.14 & 0.93 \\
 & session map & far & 0.91 & 0.94 & 0.30 & 0.66 \\
\addlinespace[2pt]
LoCoMo, MobileCLIP2 & same top-10 retrieved & near & 1.00 & 0.88 & 0.99 & -0.25 \\
 & same memories deleted & far & 0.88 & 0.96 & 0.19 & 0.84 \\
 & session map & far & 0.86 & 0.92 & 0.22 & 0.67 \\
\addlinespace[2pt]
LoCoMo, RoBERTa & same top-10 retrieved & near & 0.97 & 0.89 & 0.87 & -0.36 \\
 & same memories deleted & far & 0.92 & 0.98 & -0.03 & 0.83 \\
 & session map & far & 0.93 & 0.97 & 0.26 & 0.71 \\
\addlinespace[2pt]
LoCoMo, GloVe & same top-10 retrieved & near & 0.99 & 0.85 & 0.97 & -0.57 \\
 & same memories deleted & far & 0.92 & 0.96 & 0.43 & 0.78 \\
 & session map & far & 0.90 & 0.99 & 0.20 & 0.93 \\
\addlinespace[2pt]
20 Newsgroups, MobileCLIP2 & label precision of 10 neighbors & near & 0.59 & 0.29 & 0.68 & -0.48 \\
 & same memories deleted & far & 0.90 & 0.95 & 0.60 & 0.80 \\
 & same themes (ARI) & far & 0.77 & 0.81 & 0.25 & 0.48 \\
 & topic map & far & 0.68 & 0.79 & 0.03 & 0.56 \\
\addlinespace[2pt]
20 Newsgroups, RoBERTa & label precision of 10 neighbors & near & 0.62 & 0.41 & 0.76 & -0.65 \\
 & same memories deleted & far & 0.92 & 0.98 & -0.06 & 0.85 \\
 & same themes (ARI) & far & 0.91 & 0.88 & 0.53 & 0.12 \\
 & topic map & far & 0.93 & 0.95 & 0.38 & 0.56 \\
\addlinespace[2pt]
20 Newsgroups, GloVe & label precision of 10 neighbors & near & 0.67 & 0.34 & 0.83 & -0.71 \\
 & same memories deleted & far & 0.93 & 0.97 & 0.59 & 0.86 \\
 & same themes (ARI) & far & 0.87 & 0.89 & 0.40 & 0.52 \\
 & topic map & far & 0.80 & 0.88 & 0.09 & 0.61 \\
\addlinespace[2pt]
AG News, MobileCLIP2 & label precision of 10 neighbors & near & 0.31 & 0.49 & -0.26 & 0.47 \\
 & same memories deleted & far & 0.86 & 0.93 & 0.32 & 0.74 \\
 & same themes (ARI) & far & 0.73 & 0.83 & 0.03 & 0.59 \\
\addlinespace[2pt]
AG News, RoBERTa & label precision of 10 neighbors & near & 0.62 & 0.46 & 0.65 & -0.51 \\
 & same memories deleted & far & 0.93 & 0.98 & 0.03 & 0.83 \\
 & same themes (ARI) & far & 0.86 & 0.85 & 0.32 & 0.21 \\
\addlinespace[2pt]
AG News, GloVe & label precision of 10 neighbors & near & 0.24 & 0.22 & 0.10 & 0.04 \\
 & same memories deleted & far & 0.86 & 0.95 & 0.42 & 0.85 \\
 & same themes (ARI) & far & 0.62 & 0.80 & -0.14 & 0.65 \\
\addlinespace[2pt]

%% file: gen/tab_rescore_rows.tex
FinQA & BGE-base (96 B) & PCA + int8 & 0.77 & 0.74 & 0.60 & 0.96 \\
 &  & PCA + int8 + rescoring & 0.99 & 0.74 & 0.91 & 0.96 \\
 &  & PQ & 0.82 & 0.73 & 0.68 & 0.88 \\
 &  & PQ + rescoring & 0.99 & 0.73 & 0.95 & 0.88 \\
 &  & OPQ & 0.88 & 0.83 & 0.78 & 0.93 \\
 &  & OPQ + rescoring & 1.00 & 0.83 & 0.99 & 0.93 \\
\addlinespace[2pt]
FinQA & E5-base (96 B) & PCA + int8 & 0.77 & 0.73 & 0.57 & 0.96 \\
 &  & PCA + int8 + rescoring & 0.99 & 0.73 & 0.89 & 0.96 \\
 &  & PQ & 0.82 & 0.73 & 0.64 & 0.89 \\
 &  & PQ + rescoring & 1.00 & 0.73 & 0.94 & 0.89 \\
 &  & OPQ & 0.88 & 0.82 & 0.76 & 0.93 \\
 &  & OPQ + rescoring & 1.00 & 0.82 & 0.99 & 0.93 \\
\addlinespace[2pt]
FinQA & BGE-large (128 B) & PCA + int8 & 0.82 & 0.77 & 0.68 & 0.97 \\
 &  & PCA + int8 + rescoring & 1.00 & 0.77 & 0.96 & 0.97 \\
 &  & PQ & 0.84 & 0.74 & 0.71 & 0.89 \\
 &  & PQ + rescoring & 1.00 & 0.74 & 0.97 & 0.89 \\
 &  & OPQ & 0.91 & 0.86 & 0.83 & 0.94 \\
 &  & OPQ + rescoring & 1.00 & 0.86 & 1.00 & 0.94 \\
\addlinespace[2pt]
LoCoMo & BGE-base (96 B) & PCA + int8 & 0.76 & 0.81 & 0.69 & 0.59 \\
 &  & PCA + int8 + rescoring & 1.00 & 0.81 & 0.99 & 0.59 \\
 &  & PQ & 0.78 & 0.79 & 0.70 & 0.65 \\
 &  & PQ + rescoring & 1.00 & 0.79 & 0.98 & 0.65 \\
 &  & OPQ & 0.87 & 0.89 & 0.81 & 0.79 \\
 &  & OPQ + rescoring & 1.00 & 0.89 & 1.00 & 0.79 \\
\addlinespace[2pt]
LoCoMo & E5-base (96 B) & PCA + int8 & 0.72 & 0.77 & 0.63 & 0.43 \\
 &  & PCA + int8 + rescoring & 0.99 & 0.77 & 0.97 & 0.43 \\
 &  & PQ & 0.77 & 0.75 & 0.67 & 0.59 \\
 &  & PQ + rescoring & 0.99 & 0.75 & 0.96 & 0.59 \\
 &  & OPQ & 0.85 & 0.85 & 0.78 & 0.70 \\
 &  & OPQ + rescoring & 1.00 & 0.85 & 0.99 & 0.70 \\
\addlinespace[2pt]
LoCoMo & BGE-large (128 B) & PCA + int8 & 0.82 & 0.85 & 0.75 & 0.68 \\
 &  & PCA + int8 + rescoring & 1.00 & 0.85 & 1.00 & 0.68 \\
 &  & PQ & 0.80 & 0.80 & 0.74 & 0.66 \\
 &  & PQ + rescoring & 1.00 & 0.80 & 0.98 & 0.66 \\
 &  & OPQ & 0.90 & 0.91 & 0.86 & 0.82 \\
 &  & OPQ + rescoring & 1.00 & 0.91 & 1.00 & 0.82 \\

%% file: gen/tab_finqa_rows.tex
float16 & 1536 & 1.00 & 1.00 & 1.00 & 0.50 & 1.00 & 1.00 & 1.00 & 1.00 \\
int8 & 768 & 0.98 & 0.97 & 0.97 & 0.50 & 0.99 & 0.99 & 0.98 & 1.00 \\
int4 & 384 & 0.96 & 0.95 & 0.94 & 0.50 & 0.99 & 0.99 & 0.97 & 1.00 \\
binary, 1 bit & 96 & 0.74 & 0.65 & 0.57 & 0.44 & 0.79 & 0.80 & 0.87 & 0.97 \\
binary, 2 bit & 192 & 0.83 & 0.65 & 0.70 & 0.48 & 0.90 & 0.83 & 0.89 & 0.97 \\
binary, 4 bit & 384 & 0.86 & 0.72 & 0.76 & 0.49 & 0.94 & 0.85 & 0.90 & 0.97 \\
binary + rescoring & 96 & 0.97 & 0.65 & 0.87 & 0.49 & 0.79 & 0.80 & 0.87 & 0.97 \\
truncation 384 & 1536 & 0.86 & 0.76 & 0.74 & 0.49 & 0.77 & 0.75 & 0.89 & 0.96 \\
truncation 192 & 768 & 0.76 & 0.62 & 0.58 & 0.46 & 0.62 & 0.59 & 0.81 & 0.90 \\
truncation 96 & 384 & 0.64 & 0.47 & 0.40 & 0.33 & 0.44 & 0.46 & 0.66 & 0.80 \\
truncation 48 & 192 & 0.47 & 0.33 & 0.22 & 0.21 & 0.22 & 0.31 & 0.49 & 0.64 \\
PCA 384 & 1536 & 0.97 & 0.97 & 0.96 & 0.49 & 1.00 & 1.00 & 0.98 & 1.00 \\
PCA 192 & 768 & 0.89 & 0.86 & 0.81 & 0.44 & 0.99 & 0.92 & 0.92 & 0.99 \\
PCA 96 & 384 & 0.77 & 0.74 & 0.60 & 0.35 & 0.96 & 0.78 & 0.84 & 0.95 \\
PCA 48 & 192 & 0.63 & 0.65 & 0.39 & 0.24 & 0.92 & 0.61 & 0.60 & 0.88 \\
random projection 384 & 1536 & 0.80 & 0.66 & 0.64 & 0.47 & 0.67 & 0.67 & 0.82 & 0.91 \\
random projection 192 & 768 & 0.73 & 0.57 & 0.51 & 0.39 & 0.59 & 0.59 & 0.72 & 0.87 \\
random projection 96 & 384 & 0.62 & 0.45 & 0.37 & 0.37 & 0.34 & 0.40 & 0.57 & 0.73 \\
random projection 48 & 192 & 0.45 & 0.35 & 0.20 & 0.18 & 0.21 & 0.33 & 0.37 & 0.67 \\
PCA 192 + int8 & 192 & 0.88 & 0.84 & 0.80 & 0.43 & 0.99 & 0.93 & 0.92 & 0.99 \\
PCA 96 + int8 & 96 & 0.77 & 0.74 & 0.60 & 0.35 & 0.96 & 0.78 & 0.84 & 0.95 \\
PCA 768 + rotation + binary & 96 & 0.77 & 0.57 & 0.62 & 0.45 & 0.85 & 0.82 & 0.88 & 0.96 \\
PCA 384 + rotation + binary & 48 & 0.72 & 0.51 & 0.52 & 0.41 & 0.79 & 0.82 & 0.88 & 0.96 \\
PQ 96 & 96 & 0.82 & 0.73 & 0.68 & 0.44 & 0.88 & 0.87 & 0.90 & 0.98 \\
PQ 48 & 48 & 0.71 & 0.62 & 0.49 & 0.33 & 0.79 & 0.78 & 0.85 & 0.95 \\
PQ 24 & 24 & 0.60 & 0.51 & 0.32 & 0.20 & 0.52 & 0.65 & 0.78 & 0.89 \\
UMAP 16 & 64 & 0.52 & 0.10 & 0.24 & 0.19 & 0.08 & 0.20 & 0.69 & 0.59 \\

%% file: t_appendix_runs.tex
\section{LAQ against OPQ by run}\label{app:runs}

\begin{table}[H]
\centering\footnotesize
\caption{LAQ raises $G$ over OPQ on every held-out memory. $G$ and $L$ of OPQ and LAQ, four splits.}\label{tab:runs}
\setlength{\tabcolsep}{3.5pt}
\input{gen/tab_runs_left.tex} \hspace{2em}
\input{gen/tab_runs_right.tex} 
\end{table}

\begin{table}[H]
\centering\small
\caption{The far hinge $\mathcal L_{\mathrm{far}}$ carries the gain in $G$. Change from the OPQ start over all 52 held-out runs, each objective trained with the same recipe and checkpoint rule; means, range of the change in $G$, and runs in which $G$ is higher.}\label{tab:abl}
\setlength{\tabcolsep}{8pt}
\begin{tabular}{@{}lcccc@{}}
\toprule
Objective & $G$ change & range & higher & $L$ change \\
\midrule
$\mathcal L_{\mathrm{rec}}$ & $.0000$ & $.0000$ to $.0000$ & 0/52 & $.0000$ \\
$\mathcal L_{\mathrm{rec}}$ + near hinge & $+.0058$ & $.0000$ to $+.0124$ & 50/52 & $+.0043$ \\
$\mathcal L_{\mathrm{rec}}+\mathcal L_{\mathrm{far}}$ & $+.0127$ & $+.0022$ to $+.0453$ & 52/52 & $+.0030$ \\
LAQ, Eq.~\eqref{eq:J} & $+.0124$ & $+.0047$ to $+.0379$ & 52/52 & $+.0039$ \\
\bottomrule
\end{tabular}
\end{table}

%% file: gen/tab_runs_left.tex
\begin{tabular}{@{}lllcccc@{}}
\toprule
Memory & Encoder & Split & $G$ OPQ & $G$ LAQ & $L$ OPQ & $L$ LAQ \\
\midrule
FinQA & BGE-base & 2026 & 0.7927 & 0.8021 & 0.8537 & 0.8566 \\
 &  & 2027 & 0.7930 & 0.8020 & 0.8579 & 0.8570 \\
 &  & 2028 & 0.7885 & 0.7951 & 0.8534 & 0.8557 \\
 &  & 2029 & 0.7958 & 0.8035 & 0.8489 & 0.8514 \\
 & E5-base & 2026 & 0.7872 & 0.7959 & 0.8413 & 0.8458 \\
 &  & 2027 & 0.7858 & 0.7952 & 0.8463 & 0.8544 \\
 &  & 2028 & 0.7781 & 0.7884 & 0.8440 & 0.8486 \\
 &  & 2029 & 0.7867 & 0.7944 & 0.8469 & 0.8450 \\
 & BGE-large & 2026 & 0.8227 & 0.8280 & 0.8819 & 0.8840 \\
 &  & 2027 & 0.8209 & 0.8261 & 0.8844 & 0.8869 \\
 &  & 2028 & 0.8185 & 0.8260 & 0.8884 & 0.8910 \\
 &  & 2029 & 0.8266 & 0.8342 & 0.8879 & 0.8880 \\
\midrule
LoCoMo & BGE-base & 2026 & 0.8667 & 0.8744 & 0.8186 & 0.8224 \\
 &  & 2027 & 0.8572 & 0.8641 & 0.8349 & 0.8358 \\
 &  & 2028 & 0.8319 & 0.8440 & 0.8148 & 0.8182 \\
 &  & 2029 & 0.8722 & 0.8777 & 0.8319 & 0.8302 \\
 & E5-base & 2026 & 0.7979 & 0.8043 & 0.7651 & 0.7671 \\
 &  & 2027 & 0.7681 & 0.7785 & 0.7859 & 0.7946 \\
 &  & 2028 & 0.7703 & 0.7807 & 0.7612 & 0.7697 \\
 &  & 2029 & 0.8072 & 0.8156 & 0.7825 & 0.7806 \\
 & BGE-large & 2026 & 0.8921 & 0.8994 & 0.8496 & 0.8574 \\
 &  & 2027 & 0.8928 & 0.8989 & 0.8813 & 0.8790 \\
 &  & 2028 & 0.8678 & 0.8759 & 0.8522 & 0.8518 \\
 &  & 2029 & 0.9128 & 0.9175 & 0.8587 & 0.8602 \\
\bottomrule
\end{tabular}%

%% file: gen/tab_runs_right.tex
\begin{tabular}{@{}lllcccc@{}}
\toprule
Memory & Encoder & Split & $G$ OPQ & $G$ LAQ & $L$ OPQ & $L$ LAQ \\
\midrule
Multi-Doc & BGE-base & 2026 & 0.8196 & 0.8306 & 0.8132 & 0.8183 \\
 &  & 2027 & 0.8140 & 0.8266 & 0.8181 & 0.8208 \\
 &  & 2028 & 0.8274 & 0.8381 & 0.8102 & 0.8122 \\
 &  & 2029 & 0.8165 & 0.8271 & 0.8183 & 0.8207 \\
 & E5-base & 2026 & 0.8154 & 0.8320 & 0.8018 & 0.8092 \\
 &  & 2027 & 0.8154 & 0.8290 & 0.8068 & 0.8122 \\
 &  & 2028 & 0.8167 & 0.8302 & 0.7944 & 0.7999 \\
 &  & 2029 & 0.8140 & 0.8282 & 0.8037 & 0.8132 \\
 & BGE-large & 2026 & 0.8552 & 0.8647 & 0.8526 & 0.8572 \\
 &  & 2027 & 0.8540 & 0.8635 & 0.8564 & 0.8601 \\
 &  & 2028 & 0.8608 & 0.8686 & 0.8462 & 0.8524 \\
 &  & 2029 & 0.8545 & 0.8617 & 0.8557 & 0.8559 \\
 & Titan V2 & 2026 & 0.6680 & 0.7025 & 0.7942 & 0.8010 \\
 &  & 2027 & 0.6434 & 0.6811 & 0.7962 & 0.8043 \\
 &  & 2028 & 0.6651 & 0.7030 & 0.7848 & 0.7958 \\
 &  & 2029 & 0.6480 & 0.6803 & 0.7852 & 0.7909 \\
\midrule
FinDER & BGE-base & 2026 & 0.8139 & 0.8303 & 0.8046 & 0.8072 \\
 &  & 2027 & 0.8142 & 0.8312 & 0.8024 & 0.8055 \\
 &  & 2028 & 0.8142 & 0.8305 & 0.8029 & 0.8069 \\
 &  & 2029 & 0.8137 & 0.8289 & 0.8049 & 0.8076 \\
 & E5-base & 2026 & 0.7982 & 0.8157 & 0.7809 & 0.7893 \\
 &  & 2027 & 0.8028 & 0.8202 & 0.7793 & 0.7854 \\
 &  & 2028 & 0.8075 & 0.8238 & 0.7812 & 0.7883 \\
 &  & 2029 & 0.8032 & 0.8200 & 0.7842 & 0.7905 \\
 & BGE-large & 2026 & 0.8456 & 0.8559 & 0.8450 & 0.8480 \\
 &  & 2027 & 0.8458 & 0.8558 & 0.8445 & 0.8494 \\
 &  & 2028 & 0.8461 & 0.8563 & 0.8433 & 0.8478 \\
 &  & 2029 & 0.8439 & 0.8567 & 0.8393 & 0.8454 \\
\bottomrule
\end{tabular}%